\documentclass{article}

\usepackage[final]{neurips_2024}

\usepackage[utf8]{inputenc} % allow utf-8 input
\usepackage[T1]{fontenc}    % use 8-bit T1 fonts
\usepackage{hyperref}       % hyperlinks
\usepackage{url}            % simple URL typesetting
\usepackage{booktabs}       % professional-quality tables
\usepackage{nicefrac}       % compact symbols for 1/2, etc.
\usepackage{microtype}      % microtypography
\usepackage{xcolor}         % colors

\usepackage{color, colortbl}
\definecolor{greyC}{RGB}{180,180,180}
\definecolor{greyL}{RGB}{235,235,235}
\definecolor{citeColor}{RGB}{0,20,115}
\hypersetup{colorlinks,linkcolor={red},citecolor={citeColor},urlcolor={citeColor}}
\usepackage{multicol}
\usepackage{multirow}
\usepackage{makecell}

\definecolor{shadecolor}{rgb}{0.92,0.92,0.92}

\usepackage{amsmath}
\usepackage{amsfonts}       % blackboard math symbols
\usepackage{amssymb}
\usepackage{mathtools}
\usepackage{amsthm}
\usepackage{algorithmic,algorithm}
\usepackage{bm}

\usepackage{soul}
\usepackage{microtype}
\usepackage{booktabs}
\usepackage{caption}
\usepackage{threeparttable}
\usepackage{subfigure}
\usepackage{dsfont}
\usepackage{enumerate}
\usepackage{amsmath,amsthm,amssymb}
\usepackage{natbib}

\usepackage{authblk}

\usepackage{wrapfig}

\usepackage{framed}
\usepackage{color}
\definecolor{shadecolor}{rgb}{0.92,0.92,0.92}

\usepackage{ulem}
\usepackage{amsmath}
\usepackage{multirow}
\usepackage[pdftex]{graphicx}
\usepackage{graphicx}

\theoremstyle{plain}

\theoremstyle{definition}

\theoremstyle{remark}

\usepackage{pifont}
\newcommand{\cmark}{\ding{51}}%
\newcommand{\xmark}{\ding{55}}%

\title{Self-Calibrated Tuning of Vision-Language Models for Out-of-Distribution Detection }

\author{\\
\vspace{-8mm}
\textbf{Geng Yu}$^{1}$ \quad \textbf{Jianing Zhu}$^{2}$ \quad \textbf{Jiangchao Yao}$^{1,3} \thanks{Correspondence to Jiangchao Yao (Sunarker@sjtu.edu.cn).}$ \quad \textbf{Bo Han}$^{2,4}$\\\vspace{2mm}
$^{1}$CMIC, Shanghai Jiao Tong University \quad
$^{2}$TMLR Group, Hong Kong Baptist University \\
$^{3}$Shanghai A I Laboratory\quad
$^{4}$RIKEN Center for Advanced Intelligence Project\\\vspace{2mm}
\texttt{\{warriors30, Sunarker\}@sjtu.edu.cn}\\
\texttt{\{csjnzhu, bhanml\}@comp.hkbu.edu.hk}
}

\begin{document}

\maketitle

\begin{abstract}
  Out-of-distribution (OOD) detection is crucial for deploying reliable machine learning models in open-world applications. Recent advances in CLIP-based OOD detection have shown promising results via regularizing prompt tuning with OOD features extracted from ID data. However, the irrelevant context mined from ID data can be spurious due to the inaccurate foreground-background decomposition, thus limiting the OOD detection performance. In this work, we propose a novel framework, namely, \textit{Self-Calibrated Tuning (SCT)}, to mitigate this problem for effective OOD detection with only the given few-shot ID data. Specifically, SCT introduces modulating factors respectively on the two components of the original learning objective. It adaptively directs the optimization process between the two tasks during training on data with different prediction uncertainty to calibrate the influence of OOD regularization, which is compatible with many prompt tuning based OOD detection methods. Extensive experiments and analyses have been conducted to characterize and demonstrate the effectiveness of the proposed SCT. The code is publicly available at: \url{https://github.com/tmlr-group/SCT}.
\end{abstract}

\section{Introduction}
The deep neural networks (DNNs) are demonstrated to be overconfident on the OOD data out of the pre-defined label space~\citep{hendrycks2016baseline}, which can induce severe problems in those safety-critical applications like autonomous driving or medical intelligence. Various explorations~\citep{liang2017enhancing,djurisic2022extremely,du2022vos,zhu2024diversified} thus have been conducted in designing scoring functions or fine-tuning methods with auxiliary outliers to improve the OOD distinguishability. Specially, with the emergence of the powerful pretrained vision-language models (VLMs)~\citep{radford2021learning}, a series of prompt tuning based methods~\citep{miyai2024locoop,tao2023non,bai2023id,ming2022impact} show impressive performance in current OOD detection benchmarks, with the regularization given only few-shot in-distribution (ID) data.

\begin{figure}
  \centering
  \includegraphics[scale=0.34]{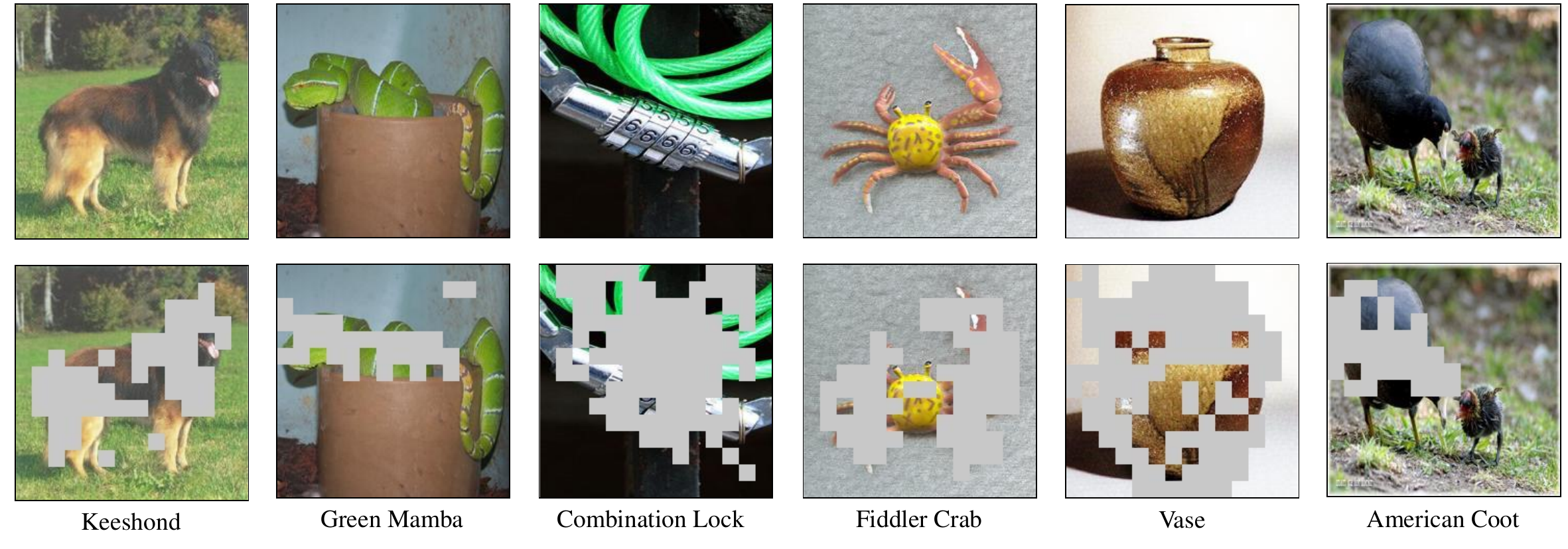}
  \hspace{0.01in}
  \vspace{1mm}
  \caption{Imperfect foreground background decomposition. The top row shows the original images from ImageNet-1k and the bottom row shows the ID-irrelevant context extracted from the original images (shown as the colored patches of images on the second row), using CLIP fine-tuned with CoOp on 16-shot data. Due to the imperfect decomposition of fine-tuned vision-language models, large portions of the extracted local features from ID data belong to ID-related regions, thus harming the performance of OOD detection.  More illustrations are presented in the Appendix~\ref{app:more_demo}.
  }
  \label{fig1:illustration}
  \vspace{-4mm}
\end{figure}

Generally, these regularizations~\citep{wang2023clipn,miyai2024locoop} are built upon the ID-irrelevant local context as the surrogate OOD source, which is extracted by VLMs (refer to Figure~\ref{fig1:illustration}) based on its alignment with ID-class text features. Although this saves the costly collection of auxiliary outliers from the open world, the quality of the ID-irrelevant local context also becomes the bottleneck, which can be greatly affected by the foreground-background decomposition with VLMs. Specifically, as revealed in previous studies~\citep{oh2023towards,tu2024empirical,wang2024open}, the prevalent VLMs struggle with poor calibration, which means that the decomposition performance on downstream data might not be well guaranteed.
Thus, it naturally motivates the following question: 
\begin{quote}
    \textit{Can we flexibly leverage the imperfect OOD features extracted by the VLM itself, to facilitate the few-shot prompt tuning for effective OOD detection}? 
\end{quote}

As illustrated in Figure~\ref{fig1:illustration}, a significant portion of extracted local context from ID data are not valid OOD features due to the inevitable imperfect decomposition. Consequently, OOD regularization based on such unreliable OOD features may potentially constrain the improvement of OOD detection. To investigate this problem, we conduct a proof-of-concept experiment on CLIP with ImageNet as the ID dataset and prompt-tune the model with different groups of data divided by their overall prediction uncertainty. Specifically, we find that the performance of prompt tuning based methods significantly deteriorates as the uncertainty of the given ID data rises, as presented in Figure~\ref{fig2:emprical_demo}, which motivates us to leverage such clues to overcome the current issue. Intuitively, as the model prediction on ID samples is less certain, the OOD features extracted from these data are less reliable. Performing OOD regularization on such unreliable surrogate OOD features can degrade the OOD detection performance of CLIP. Therefore, a potential idea is to adaptively adjust the importance of OOD features extracted from ID data according to their prediction uncertainty during model training.

Based on the previous observation, we propose a new learning framework, i.e., \textit{\underline{S}elf-\underline{C}alibrated \underline{T}uning} (SCT), to alleviate the problem induced by spurious OOD features. At the high level, we aim to dynamically adjust the weight of OOD regularization from different training samples based on their prediction uncertainty to calibrate their influence on model training. In detail, we introduce modulating factors based on the sample uncertainty estimation respectively on the two parts of the original learning objective of prompt tuning for OOD detection (refer to Eq~\eqref{eq: SCT}). Under this new learning framework, the model’s attention is directed towards the classification task to better generalize to the downstream ID dataset when training with low-confidence data. OOD features extracted from high-confidence ID data are attached more importance to achieve more effective OOD regularization. The redirection effect of these two modulating factors facilitates VLMs learning from imperfect OOD features to ultimately improve the OOD detection of prompt tuning. Our main contributions can be summarized as follows,
\begin{itemize}
    \item Conceptually, we investigate the problem of imperfect OOD features extracted in prompt tuning based OOD detection methods and observe that ID data with different prediction uncertainty exhibits a distinctive influence on the OOD regularization. (in Section~\ref{sec:motivation})
    \item Technically, we propose a novel learning framework, namely \textit{Self-Calibrated Tuning} (SCT), for facilitating prompt tuning for effective OOD detection given only few-shot ID samples, which conducts adaptive redirection of model optimization process between the two tasks to calibrate the influence of OOD features mined from different ID data. (in Sections~\ref{sec:SCT})
    \item Empirically, extensive experiments from different perspectives are conducted to verify the effectiveness of SCT in improving OOD detection performance. To be specific, SCT improves the false positive rate (FPR95) by 3\% compared to the previous best method on the large-scale ImageNet-1k~\citep{deng2009imagenet} benchmark. Furthermore, we perform various ablations and further discussions to provide a thorough understanding. (in Section~\ref{exp})

    %we conduct extensive explorations to verify the overall effectiveness of our method in improving OOD detection performance, and perform various ablations to provide a thorough understanding. %Using various ID and OOD benchmarks, we provide comprehensive results across different setups and further discussion.
\end{itemize}

\section{Related works}

\textbf{Prompt tuning for VLMs.} The concept of prompt tuning was originally applied in the field of natural language processing~\citep{radford2018improving}. To eliminate the need for manual prompt crafting, prompt tuning exploits supervision signals from downstream tasks to automate the process of prompt generation. Autoprompt~\citep{shin2020autoprompt} searches for tokens that cause the greatest changes in gradients based on the label likelihood. Prefix-Tuning~\citep{li2021prefix} introduces a sequence of continuous vectors that can be end-to-end optimized in the token embedding space. Incorporating prompt tuning into computer vision. CoOp~\citep{zhou2021learning} adapts pretrained vision-language models by optimizing a set of learnable continuous prompt vectors.  Various prompt tuning methods~\citep{zhou2022cocoop,khattak2023maple,sun2022dualcoop} have been subsequently proposed to address different vision tasks. However, since these methods are not developed for OOD detection, they face challenges in identifying unknown OOD samples encountered at inference stages.

\textbf{Out-of-distribution detection with VLMs.} Large pretrained vision-language models have enriched the landscape of OOD detection through their remarkable generalization capability in both visual and textual domains. MCM~\citep{ming2022delving} employs the concept of maximum softmax probability \citep{hendrycks2016baseline} into the inference process of CLIP for OOD detection, while CLIPN~\citep{wang2023clipn} trains an additional text encoder using and set of prompts with a large external dataset to improve its negative semantic understanding. Compared with zero-shot methods, prompt tuning based approaches achieve better OOD detection with access to few-shot ID training data. LoCoOp~\citep{miyai2024locoop} adopts prompt tuning and extracts ID-irrelevant background from CLIP’s local features as surrogate OOD data to regularize the learned prompts. ~\citep{bai2023id} discovers ID-like outliers from ID samples via random cropping to learn a set of negative prompts. However, these two representative prompt learning-based methods suffer from spurious OOD features extracted from ID data due to the imperfect foreground-background decomposition of VLMs. In addition, LSN~\citep{nie2023out} introduce negative prompts to empower VLMs to learn negative semantics from ID samples while NegPrompt~\citep{li2024learning} leverages negative prompts to investigate the novel setting of open-vocabulary OOD detection.

\section{Method}
\label{headings}

In this section, we introduce our new framework, i.e., \textit{Self-Calibrated Tuning} (SCT), which conducts adaptive redirection of model learning far away from OOD data region during prompt tuning with only the few-shot ID data. Firstly, we provide preliminaries and notations about prompt tuning based OOD detection (Section~\ref{sec:prelimi}). Secondly, we present and discuss the critical motivation that inspires our method (Section~\ref{sec:motivation}). Thirdly, we introduce its newly derived learning objective with the explanation and analysis of the underlying intuition and present its algorithmic realization (Section~\ref{sec:SCT}).

\subsection{Preliminaries}
\label{sec:prelimi}

VLM-based OOD detection aims to identify test samples that do not belong to any ID class designated by the downstream tasks~\citep{miyai2024generalized}. Therefore, the ID distribution is defined by the ID classes from the downstream tasks, which are different from those of the upstream pretraining. Formally, we consider multi-class classification as the original training task~\citep{nguyen2015deep}, where $\mathcal{X}\subset\mathbb{R}^d$ denotes the input space and $\mathcal{Y}=\{1,\ldots, M\}$ denotes the label space. A reliable classifier should be able to detect the OOD input, which can be considered a binary classification problem. We consider $\mathcal{D}_\text{in}$ as the distribution of ID data over pairs of examples $\boldsymbol{x} \in \mathbb{R}^d$ and corresponding labels $y \in \mathcal{Y}$. At test time, the environment can present a distribution $\mathcal{D}_\text{out}$ over $\mathcal{X}$ of OOD data. In general, the OOD distribution $\mathcal{D}_\text{out}$ is defined as an irrelevant distribution of which the label set has no intersection with $\mathcal{Y}$~\citep{zhu2023unleashing} and thus should not be predicted by the model parameterized by $\theta$. A decision model $\Gamma(\cdot)$ can be made with the threshold $\mu$:

\begin{equation}
    \Gamma_{\mu}(\boldsymbol{x};\theta)=\left \{
    \begin{aligned}
    &\text{ID} && S(\boldsymbol{x}; \theta)\geq\mu\\
    &\text{OOD} && S(\boldsymbol{x}; \theta)<\mu
    \end{aligned},
    \right.
\end{equation}

\textbf{Vanilla prompt tuning}. Given an ID image $\boldsymbol{x}$ and its corresponding label $y$, a global visual feature $\boldsymbol{f} = f(\boldsymbol{x})$ is obtained by the visual encoder $f$ of CLIP. Then, the textual prompt vectors can be formulated as $\boldsymbol{t}_m = \{\boldsymbol{\omega}_1, \boldsymbol{\omega}_2, . . . , \boldsymbol{\omega}_N, \boldsymbol{c}_m\}$, where $\boldsymbol{c}_m$ denotes the word embedding of the ID class name and $\boldsymbol{\omega}= \{\boldsymbol{\omega}_n|_{n=1}^N\}$ are $N$ learnable context vectors, each of which has the same dimension as the word embedding. The text encoder $g$ takes prompt $\boldsymbol{t}_m$ as the input and outputs the textual feature as $\boldsymbol{g}_m = g(\boldsymbol{t}_m)$. The final prediction probability of the CLIP model is computed as follows:
\small
\begin{equation}
\begin{aligned}
p(y=m \mid \boldsymbol{x}; \boldsymbol{\omega})=\frac{\exp \left(\operatorname{sim}\left(\boldsymbol{f}, \boldsymbol{g}_m\right) / \tau\right)}{\sum_{m^{\prime}=1}^M \exp \left(\operatorname{sim}\left(\boldsymbol{f}, \boldsymbol{g}_{m^{\prime}}\right) / \tau\right)},
\end{aligned}
\label{eq:prob_coop}
\end{equation}
\normalsize
where $\operatorname{sim}\left(\cdot, \cdot \right)$ denotes cosine similarity, and $\tau$ represents the temperature of Softmax. We use $\boldsymbol{p}(\boldsymbol{x}; \boldsymbol{\omega})$ to represent the probability vector $[p(y=1|\boldsymbol{x}; \boldsymbol{\omega}), p(y=2|\boldsymbol{x}; \boldsymbol{\omega}), \dots, p(y=M|\boldsymbol{x}; \boldsymbol{\omega})]$, denoting the prediction probability of ID image $\boldsymbol{x}$ for every ID class. 

\textbf{Prompt tuning for OOD detection}. Compared with vanilla prompt tuning methods like CoOp~\citep{zhou2021learning}, advanced prompt tuning based OOD detection methods extract surrogate OOD features from ID data via various methods to perform OOD regularization. LoCoOp~\citep{miyai2024locoop} can further improve the detection performance by regularizing the learnable prompts $\boldsymbol{\omega}$ on OOD features $\boldsymbol{\tilde{X}}$ extracted from ID local features using a ranking-based method, and its corresponding learning objective is defined as follows,
\begin{equation}
    {\mathcal{L}}_\text{LoCoOp}=\mathbb{E}_{(\boldsymbol{x},y) \sim \mathcal{D}_\text{in}}\left[\ell_\text{CE}(p(y|\boldsymbol{x}; \boldsymbol{\omega}),y) + \lambda \ell_\text{OOD}(\boldsymbol{p}(\boldsymbol{\tilde{X}};\boldsymbol{\omega}))\right], \label{eq: locoop}
\end{equation}
where $\lambda$ is the balancing parameter,  $\ell_\text{CE}(\cdot)$ is the Cross-Entropy (CE) loss, $\boldsymbol{p}(\boldsymbol{\tilde{X}})$ is the prediction probability vector for $\boldsymbol{\tilde{X}}$ and $\ell_\text{OOD}(\cdot)$ is the negative entropy of the given probability vector. 
IDLike~\citep{bai2023id} conducts multiple random cropping on ID data and chooses cropped regions as OOD features based on their feature similarity with textual features of ID classes.

%$\ell_\text{OOD}(\boldsymbol{p}(\boldsymbol{\tilde{X}};\boldsymbol{\omega}))$ can be written as $\sum_{m^{\prime}=1}^M p_{m^{\prime}}(\boldsymbol{\tilde{X};\boldsymbol{\omega}}) \log p_{m^{\prime}}(\boldsymbol{\tilde{X};\boldsymbol{\omega}})$. 

\begin{figure}
  \centering
  \includegraphics[width=9cm,height=4.6cm]{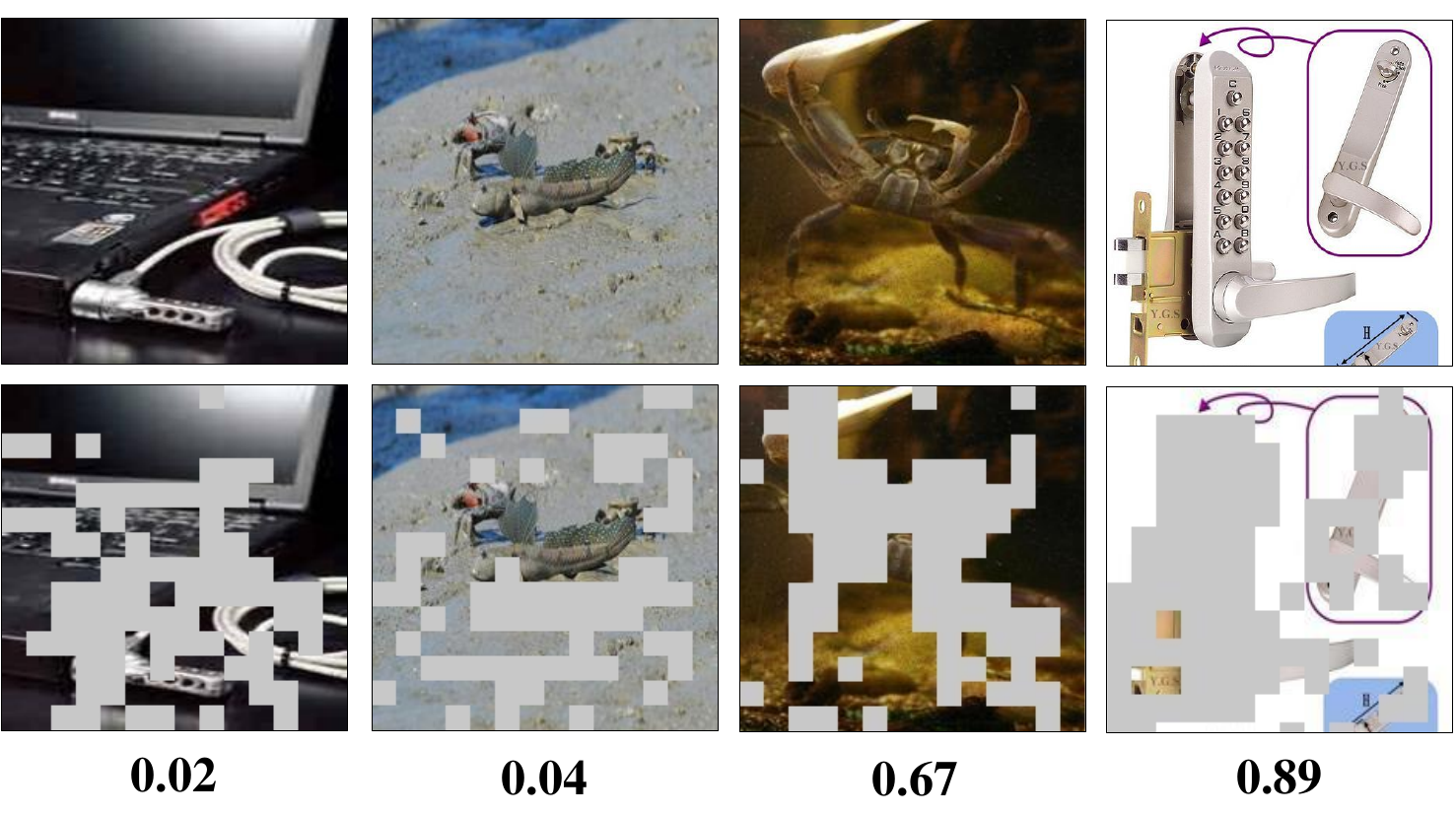}
  \includegraphics[scale=0.20]{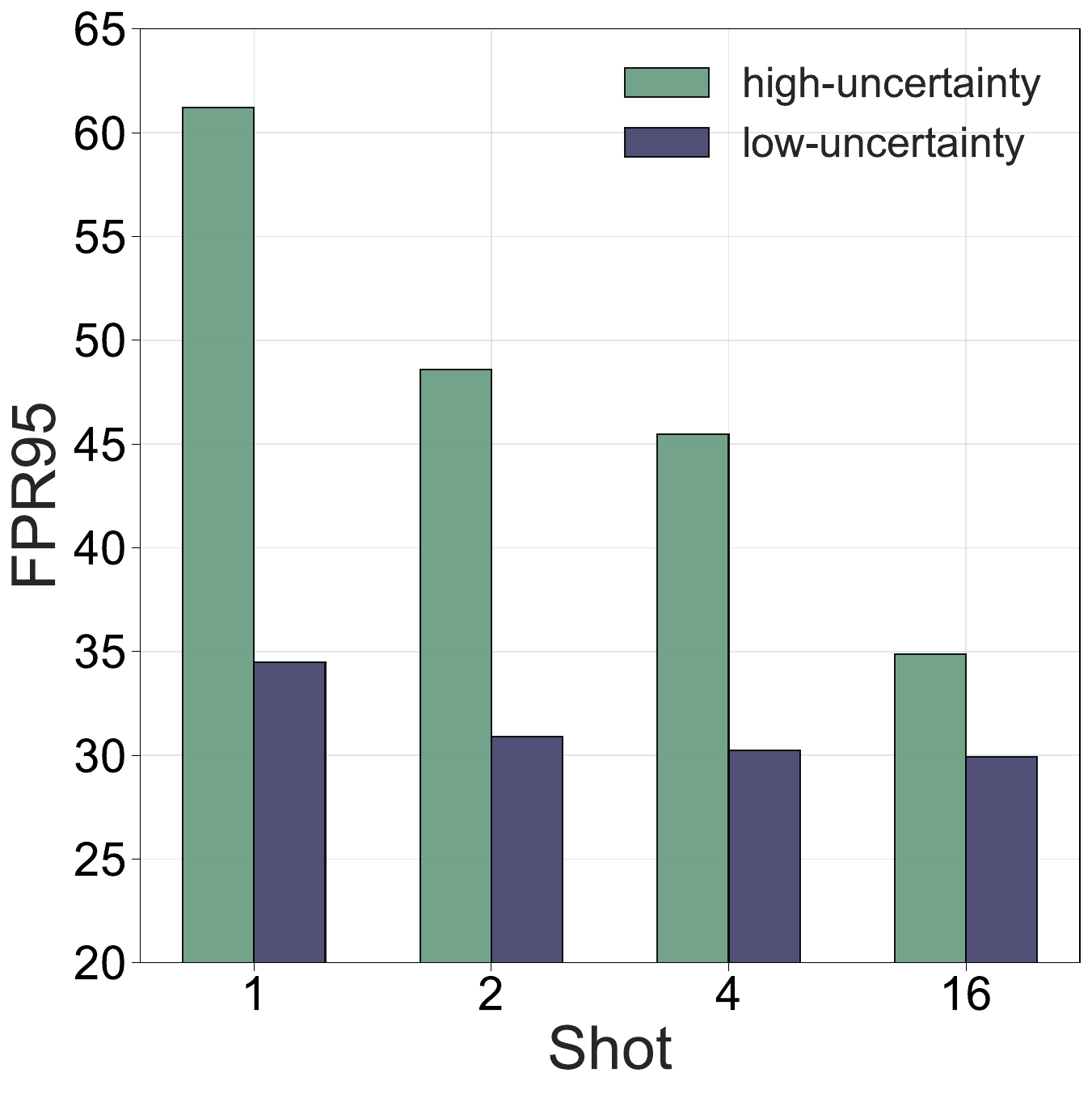}
  \caption{Empirical demonstration about invalid OOD features extracted from ID data in LoCoOp and the influence of sample uncertainty on OOD detection performance. In the left and middle panels, we illustrate the extracted OOD features at different levels of uncertainty and find that they become unreliable as the uncertainty increases. The numbers at the bottom denote the prediction probability for the ground-truth labels from fine-tuned CLIP. In the right panel, we collect ID samples of different uncertainty levels based on prompt-tuned CLIP and divide them into 2 groups. The result demonstrates that the OOD detection performance of LoCoOp significantly degrades as the uncertainty level of ID data rises. We leave the experimental details in Appendix~\ref{app:add_exp_setup} for reference.}
  \label{fig2:emprical_demo}
  \vspace{-4mm}
\end{figure}

\subsection{Motivation}
\label{sec:motivation}

Given the only ID data, previous studies propose to extract the ID-irrelevant local context as the surrogate OOD source, which depends on the foreground-background decomposition using CLIP. However, since the model itself has uncertainty on the prediction results, the correctness of the decomposition cannot always be guaranteed. Thus, as illustrated in Figure~\ref{fig1:illustration}, the extracted local regions based on the prediction of CLIP may result in spurious OOD features, which then limits the OOD detection performance. Thus, it naturally motivates the following critical research question:  

\begin{quote}
\textit{How could we better utilize the surrogate OOD features extracted by imperfect foreground-background decomposition of CLIP for effective OOD regularization?}
\end{quote}

In this work, we conduct a proof-of-concept experiment to investigate the relationship between sample uncertainty and OOD detection performance of prompt tuning based methods. We use ViT-B/16 CLIP as the model and large-scale ImageNet-1k OOD benchmarks.
 First, we observe that the quality of OOD features extracted by CLIP is highly correlated with the uncertainty estimation of ID data. The extracted OOD features become more spurious as the uncertainty escalates, as illustrated in the left and middle panel of Figure~\ref{fig2:emprical_demo}. Furthermore, as empirically shown in the right panel of Figure~\ref{fig2:emprical_demo}, the performance of prompt tuning based methods is heavily affected by the overall uncertainty level of the given ID data. Intuitively, as the predictions of the model on ID samples are less confident, the OOD features extracted from these data are less reliable. Regularizing on such invalid OOD features can undermine the calibration ability and OOD detection performance of CLIP, which can be reflected by the higher FPR95 score (indicating a higher error on OOD detection). 
 Therefore, a new mechanism is required to take the sample uncertainty into consideration to assist the model in learning from these imperfect OOD features for more effective OOD detection.

\subsection{Self-calibrated tuning}
\label{sec:SCT}
As aforementioned, the LoCoOp-based OOD detection paradigm relies on the extracted ID-background local features for OOD regularization.  Given the imperfect foreground-background decomposition, the model is expected to effectively learn from the inaccurate OOD features for better OOD detection. Inspired by the previous observation as shown in Section \ref{sec:motivation}, one conceptual idea to mitigate this problem is to adaptively adjust the importance of OOD regularization generated from different ID samples based on uncertainty estimation to alleviate the wrong guidance of invalid OOD features. Under this learning paradigm, the model can be regularized by more valid OOD features and simultaneously prevent itself from overconfidence to improve OOD detection. To this intuition, we consider reformulating the learning objective under the framework of prompt tuning as follows,
\begin{equation}
    {\mathcal{L}}_\text{SCT}=\mathbb{E}_{(\boldsymbol{x},y) \sim \mathcal{D}_\text{in}}\left[\ell_\text{CE}(p(y|\boldsymbol{x};\boldsymbol{\omega}),y) * \phi(p(y|\boldsymbol{x};\boldsymbol{\omega})) + \lambda \ell_\text{OOD}(\boldsymbol{p}(\tilde{X});\boldsymbol{\omega}) * \psi(p(y|\boldsymbol{x};\boldsymbol{\omega}))\right], \label{eq: SCT}
\end{equation}
where $\phi:\mathbb{R}^{M} \rightarrow \mathbb{R}$ and $\psi:\mathbb{R}^{M} \rightarrow \mathbb{R}$ indicate the newly-introduced modulating functions that calculate adaptive factors for the two components of the original loss function (i.e., Eq~\eqref{eq: locoop}) of LoCoOp based on the uncertainty estimation. In this loss function, the left part is for the ID classification task, and the right part is for the OOD regularization. Specifically, $\phi$ should be monotonically decreasing and $\psi$ should be monotonically increasing with respect to $p(y|\boldsymbol{x};\boldsymbol{\omega})$ so that the modulating factors shift the focus of prompt learning between the two tasks during the training process. 

In detail, when the model outputs low-confidence prediction for the ground-truth label, the importance of the ID classification task is highlighted in order to better generalize to the downstream task and simultaneously reduce the effect of regularization from invalid OOD features extracted from ID data. When the model can accurately and confidently classify the ID samples, its attention is redirected towards OOD regularization to strengthen the positive effect of useful ID-irrelevant features for better OOD detection. In the meantime, the loss contribution of the classification task is reduced to avoid the model overfitting to the downstream dataset, which benefits the calibration of the model~\citep{mukhoti2020calibrating} and further enhances the validity of extracted OOD features. 

Under this learning framework, the goal of confidence calibration and OOD detection can benefit from each other. This method calibrates the influence of OOD features mined from different ID data based on model prediction confidence during the training process to facilitate capturing more reliable OOD features from ID data. Among a wide range of functions that satisfy the simple requirements as discussed above, we choose the linear function due to its simple design. Concretely, we formulate the loss function of SCT as follows,

\begin{equation}
    {\mathcal{L}}_\text{SCT}=\mathbb{E}_{(\boldsymbol{x},y) \sim \mathcal{D}_\text{in}}\left[\ell_\text{CE}(p(y|\boldsymbol{x};\boldsymbol{\omega}),y) * (1-p(y|\boldsymbol{x};\boldsymbol{\omega})) + \lambda \ell_\text{OOD}(\boldsymbol{p}(\tilde{X};\boldsymbol{\omega})) * p(y|\boldsymbol{x};\boldsymbol{\omega})\right], \label{eq: SCT_example}
\end{equation}

This implementation of ${\mathcal{L}}_\text{SCT}$ introduces no extra hyperparameters and we empirically demonstrate its effectiveness in the following experiment section~\ref{exp}. In the Appendix~\ref{app:add_exp_result}, we consider other instantiations of the modulating function and demonstrate that these can be comparably effective. 

\paragraph{Extraction of OOD local features.}
We adopt the ranking-based method as suggested by LoCoOp~\citep{miyai2024locoop} to extract OOD local features from ID samples for OOD regularization. To be specific, we select the region indices of ID-irrelevant regions from a set of all region indices $I = \{0, 1, 2, ..., H\times W-1\}$, where $H$ and $W$ denote the height and width of the feature map. We calculate the classification prediction probabilities for each region $i$ during training by computing the similarity between the image features $\boldsymbol{f}^{(i)}$ of each region $i$ and the text features of the ID classes. Then we choose regions that do not include their ground truth class in the top-$K$ predicted classes as ID-irrelevant regions $J$.
The corresponding formulation is presented as follows:
\begin{equation}
\begin{aligned}
J = \{i \in I : \operatorname{rank}(p^{(i)}(y|\boldsymbol{x};\boldsymbol{\omega})) > K\},
\end{aligned}
\label{eq:extract_ood}
\end{equation}
\normalsize
where $p^{(i)}(y|\boldsymbol{x};\boldsymbol{\omega})$ denotes the prediction probability of region $i$ for the ground-truth label and $\operatorname{rank}(p^{(i)}(y|\boldsymbol{x};\boldsymbol{\omega}))$ denotes the rank of the ground-truth label among all ID classes. We summarize the whole procedure of the proposed SCT in Algorithm~\ref{alg:SCT}.
% , which consists of multi-round training iterations.

\begin{algorithm}[t!]
%\footnotesize
   \caption{Self-Calibrated Tuning(SCT)}
   \label{alg:SCT}
   {\bf Input:} learnable prompt: $\boldsymbol{\omega}$, fine-tuning epochs: $T$, learning rate: $\eta$, a training set of ID data, extraction rank parameter: $K$, regularization term weight: $\lambda$;\\
   {\bf Output:} fine-tuned prompt $\boldsymbol{\omega}$;
\begin{algorithmic}[1]
    %\STATE \colorbox{shadecolor}{Initialize a popup score for every weight}
    %\STATE \begin{footnotesize}\texttt{// Comment}\end{footnotesize}\vspace{2mm}
    \FOR{epoch $= 1$, $\dots$, $T$}
    \FOR {mini-batch $=1$, $\dots$, $M$}
    \STATE Sample a mini-batch $\{(\boldsymbol{x}_i, y_i) \}^{n}_{i=1}$ from the training set
    \STATE Sample surrogate OOD features $\tilde{X}$ from ID local feature map by Eq.~\eqref{eq:extract_ood}
    \STATE $\boldsymbol{\omega} \gets \boldsymbol{\omega} - \eta \nabla_{\boldsymbol{\omega}} \bigg \{ \frac{1}{n}\sum\ell_\text{CE}(p(y|\boldsymbol{x};\boldsymbol{\omega}),y) * \phi(p(y|\boldsymbol{x};\boldsymbol{\omega}))
     + \lambda \ell_\text{OOD}(\boldsymbol{p}(\tilde{X});\boldsymbol{\omega}) * p(y|\boldsymbol{x};\boldsymbol{\omega})  \bigg \}$
  \ENDFOR
 \ENDFOR
\end{algorithmic}
\end{algorithm}

\paragraph{Test-time OOD detection.}
At the testing stage, we use the GL-MCM score proposed by~\citep{miyai2023zero} since it has been empirically proved to outperform the conventional MCM score. It combines the maximum softmax probability scores for both global and local image features.
The detailed formulation is presented as follows:
\begin{equation}
\label{eq:gl_mcm}
S_{\mathrm{GL-MCM}} = \max _{m} \frac{\exp \left(\operatorname{sim}\left(\boldsymbol{f}, \boldsymbol{g}_m\right) / \tau\right)}{\sum_{m^{\prime}=1}^M \exp \left(\operatorname{sim}\left(\boldsymbol{f}, \boldsymbol{g}_{m^{\prime}}\right) / \tau\right)} + \max _{m, i} \frac{\exp \left(\operatorname{sim}\left(\boldsymbol{f}^{(i)}, \boldsymbol{g}_m\right) / \tau\right)}{\sum_{m^{\prime}=1}^M \exp \left(\operatorname{sim}\left(\boldsymbol{f}^{(i)}, \boldsymbol{g}_{m^{\prime}}\right) / \tau\right)}. 
\end{equation}
where $\boldsymbol{f}^{(i)}$ denotes the local image feature of ID samples $\boldsymbol{x}$ for region $i$ and we set $\tau = 1$.

% \paragraph{Theorectical Analysis.}

\paragraph{Comparison and compatibility.}  Compared with the previous prompt tuning based OOD detection algorithm~\citep{miyai2024locoop}, the critical idea behind SCT is trying to adaptively adjust the contribution of OOD features extracted from ID data with different uncertainty during training. It provides a general framework (under the guidance of the learning objective in Eq.~\eqref{eq: SCT}) to assist VLMs in learning from spurious OOD features for effective OOD detection. The discriminative feature regularized by our SCT can be utilized by those advanced post-hoc scoring functions~\citep{sun2021react,huang2021importance,sun2022dice}. For prompt tuning based methods, the adaptive modulation introduced in SCT is orthogonal to current tuning objectives~\citep {liu2020energy} and also compatible with different augmentation~\citep{lu2023likelihood} or mining strategies~\citep{bai2023id,zhu2024diversified}.

\section{Experiment}
\label{exp}

In this section, we present the comprehensive verification of the proposed SCT in the CLIP-based OOD detection scenario. First, we provide the experimental setups in detail (in Section~\ref{sec:setup}). Secondly, we provide the performance comparison of our approach with a series of CLIP-based post-hoc methods and prompt tuning based methods (in Section~\ref{sec:main_results}). Thirdly, we conduct various ablation studies and further discussions to understand our method (in Section~\ref{sec:ablation}).

\subsection{Experimental setup}
\label{sec:setup}
\textbf{Datasets.}
Following the common benchmarks used in previous works, we adopt the ImageNet-1K dataset~\citep{deng2009imagenet} as the ID data. For OOD datasets, we adopt the same ones as in~\citep{huang2021mos}, including subsets of iNaturalist~\citep{van2018inaturalist}, SUN~\citep{xiao2010sun}, Places~\citep{zhou2017places}, and TEXTURE~\citep{cimpoi2014describing}. For the few-shot training, we use 1, 2, 4, and 16 shots ID data for training, respectively, and evaluate models in the full test set. We also present the comparison results on conventional CIFAR benchmarks~\citep{hendrycks2018deep,liu2020energy}, which adopt CIFAR-10 and CIFAR-100 as ID datasets~\citep{krizhevsky2009learning_cifar10}, in Appendix~\ref{app:add_exp_result}. 

\textbf{Evaluation metrics.} We employ the following two common metrics to evaluate the performance of OOD detection: (i) Area Under the Receiver Operating Characteristic curve (AUROC) \citep{davis2006relationship} can be interpreted as the probability for a positive sample to have a higher discriminating score than a negative sample \citep{fawcett2006introduction}; (ii) False Positive Rate (FPR) at $95\%$ True Positive Rate (TPR) \citep{liang2017enhancing} indicates the probability for a negative sample to be misclassified as positive when the true positive rate is at $95$\%. We also adopt in-distribution testing accuracy (ID-ACC) to measure the preservation level of the performance for the original classification task on ID data and use Expected Calibration Error (ECE)~\citep{naeini2015obtaining} to measure the performance of SCT on confidence calibration, which are both presented in Appendix~\ref{app:add_exp_result}.

\textbf{Implementation details.} Following previous works~\citep{tao2023non, ming2022delving, miyai2023zero}, we use ViT-B/16~\citep{dosovitskiy2020image} as the backbone model for the main experiment. For the hyperparameter $K$ in the surrogate OOD features extraction, we use 200 in all experiments as recommended by~\citep{miyai2024locoop}.  For SCT, we adopt $\lambda=0.4$ under the 1-shot setting and $\lambda=0.2$ under the 16-shot setting. We train the CLIP for 25 epochs with a learning rate of 0.002 and other hyperparameters (\textit{e.g.} batch size=32, SGD optimizer and token lengths $N$=16) are the same as those of CoOp~\citep{zhou2021learning}. We use two Nvidia 3090 GPUs for all experiments.

\textbf{OOD detection baselines.} We compare SCT with several competitive CLIP-based OOD detection methods in the two directions, including post-hoc methods and prompt tuning based methods. For post-hoc methods, we compare with MCM~\citep{ming2022delving} and GL-MCM~\citep{miyai2024locoop} as zero-shot baselines. We also adopt Maximum Softmax Probability (MSP) \citep{hendrycks2016baseline}, ODIN \citep{liang2017enhancing}, ReAct~\citep{sun2021react}, MaxLogit~\citep{hendrycks2019scaling}, and Energy score \citep{liu2020energy} as conventional scoring function baselines. In addition, we provide more discussions about post-hoc methods and our methods in Appendix~\ref{app:add_exp_result}. For prompt tuning based methods, we adopt CoOp~\citep{zhou2021learning}, LoCoOp~\citep{miyai2024locoop}, IDLike \citep{bai2023id}, NegPrompt~\citep{li2024learning} and LSN~\citep{nie2023out} as baselines. For all prompt tuning based methods, we constrain all major experiments to few-shot learning scenarios, which is more practical in real cases. Note that LSN is a general learning framework that can be combined with various prompt tuning methods and we report the result of LSN incorporated with LoCoOp in Table~\ref{tab:main_table1}. We leave more definitions and implementation details in the Appendix~\ref{app:baseline_info}.

\begin{table*}[t!]
    \caption{\textbf{Comparison results on ImageNet-1k OOD benchmarks.} All methods are trained on the same backbone CLIP-ViT-B/16. Bold numbers are superior results. $\uparrow$ indicates larger values are better, and $\downarrow$ indicates smaller values are better. Results marked with $\dag$ are taken from ~\citep{wang2023clipn} and ~\citep{miyai2024locoop}. The prompt tuning based methods are run under multiple trials with reporting the mean and standard deviation of the performance.}
    \vspace{2mm}
    \centering
    \renewcommand\arraystretch{1.0}
    \resizebox{\textwidth}{!}{
    \begin{tabular}{lllllllllll}
        \toprule[1.5pt]
         \multirow{3}*{\textbf{Method}} & \multicolumn{8}{c|}{\textbf{OOD dataset}}\\
        %\cmidrule{3-8}
         ~ & \multicolumn{2}{c}{\textbf{iNaturalist}} & \multicolumn{2}{c}{\textbf{SUN}} & \multicolumn{2}{c}{\textbf{Places365}} &  \multicolumn{2}{c}{\textbf{Textures}} & \multicolumn{2}{c}{\textbf{Average}}\\
         ~ & FPR95$\downarrow$ & AUROC$\uparrow$ & FPR95$\downarrow$ & AUROC$\uparrow$ & FPR95$\downarrow$ & AUROC$\uparrow$  & FPR95$\downarrow$ & AUROC$\uparrow$  & FPR95$\downarrow$ & AUROC$\uparrow$\\
        \midrule[0.6pt]
        ~ & \multicolumn{10}{c}{{\textit{Zero-shot methods}}} \\
        MCM & 31.86 & 94.17 & 37.28 & 92.55 & 42.94 & 90.09 & 58.37 & 85.83 & 42.61 & 90.66 \\ 
        GL-MCM & 15.16 & 96.71 & 29.16 & 93.41 & 37.07 & 90.37 & 58.85 & 83.11 & 35.06 & 90.90\\
        \midrule[0.6pt]
        ~ & \multicolumn{10}{c}{\textit{CLIP-based post-hoc methods}} \\
        MSP$^\dag$ & 74.57 & 77.74 & 76.95 & 73.97 & 79.72 & 72.18 & 73.66 & 74.84 & 74.98 & 76.22 \\
        ODIN $^\dag$ & 98.93 & 57.73 & 88.72 & 78.42 & 87.80 & 76.88 & 85.47 & 71.49 & 90.23 & 71.13\\
        Energy$^\dag$ & 64.98 & 87.18 & 46.42 & 91.17 & 57.40 & 87.33 & 50.39 & 88.22  & 54.80 & 88.48\\
        ReAct$^\dag$ & 65.57 & 86.87 & 46.17 & 91.04 & 56.85 & 87.42 & 49.88 & 88.13 & 54.62 & 88.37 \\
        MaxLogit$^\dag$ & 60.88 & 88.03 & 44.83 & 91.16 & 55.54 & 87.45 & 48.72 & 88.63 & 52.49 & 88.82 \\
        \midrule[0.6pt]
        ~ & \multicolumn{10}{c}{\textit{Prompt tuning based methods}} \\
        ~ & \multicolumn{10}{c}{1-shot}\\
         CoOp & $43.80^{\pm4.33}$ & $91.40^{\pm0.70}$ & $35.42^{\pm1.56}$ & $92.65^{\pm0.53}$ & $40.70^{\pm2.63}$ & $90.49^{\pm0.68}$ & $49.61^{\pm3.03}$ & $87.95^{\pm0.79}$ & $42.38^{\pm1.92}$ & $90.63^{\pm0.38}$ \\
         LoCoOp & $28.81^{\pm2.78}$ & $94.05^{\pm0.72}$ & $25.76^{\pm0.53}$ & $94.51^{\pm0.19}$ & $33.68^{\pm0.62}$ & $91.59^{\pm0.23}$ & $51.53^{\pm0.42}$ & $86.85^{\pm0.24}$ & $34.94^{\pm0.65}$ & $91.75^{\pm0.23}$ \\
         IDLike & $12.07^{\pm0.88}$ & $97.65^{\pm0.10}$ & $40.55^{\pm5.84}$ & $91.07^{\pm1.80}$ & $47.94^{\pm5.24}$ & $88.31^{\pm2.05}$ & $38.34^{\pm13.39}$ & $89.67^{\pm4.03}$ & $34.72^{\pm0.80}$ & $91.67^{\pm0.07}$ \\
         NegPrompt & $65.03^{\pm8.69}$ & $84.56^{\pm2.52}$ & $44.39^{\pm1.66}$ & $89.63^{\pm0.66}$ & $51.31^{\pm6.21}$ & $86.55^{\pm2.19}$ & $87.60^{\pm1.61}$ & $63.76^{\pm3.02}$ & $62.08^{\pm3.71}$ & $81.13^{\pm1.78}$ \\
         LSN  & $59.28^{\pm7.02}$ & $87.20^{\pm3.15}$ & $40.15^{\pm0.82}$ & $91.47^{\pm0.14}$ & $46.11^{\pm1.86}$ & $88.74^{\pm0.57}$ & $60.34^{\pm0.14}$ & $83.92^{\pm0.42}$ & $51.47^{\pm1.53}$ & $87.84^{\pm0.58}$ \\
          SCT & $19.16^{\pm1.15}$ & $95.70^{\pm0.28}$ & $23.52^{\pm1.91}$ & $94.58^{\pm0.56}$ & $32.81^{\pm1.14}$ & $91.23^{\pm0.32}$ & $48.87^{\pm1.38}$ & $86.66^{\pm0.33}$ & $\textbf{31.09}^{\pm0.48}$ & $\textbf{92.04}^{\pm0.25}$ \\
         \midrule[0.6pt]
         ~& \multicolumn{10}{c}{16-shot}\\
         CoOp & $28.25^{\pm4.68}$ & $93.92^{\pm1.17}$ & $31.15^{\pm0.89}$ & $93.13^{\pm0.38}$ & $39.12^{\pm1.05}$ & $90.50^{\pm0.37}$ & $41.86^{\pm1.88}$ & $90.40^{\pm0.54}$ & $35.09^{\pm1.60}$ & $91.99^{\pm0.42}$ \\
         LoCoOp & $17.58^{\pm2.22}$ & $96.30^{\pm0.57}$ & $22.82^{\pm0.34}$ & $95.20^{\pm0.06}$ & $32.21^{\pm0.53}$ & $92.03^{\pm0.20}$ & $45.27^{\pm0.95}$ & $88.86^{\pm0.26}$ & $29.47^{\pm0.29}$ & $93.10^{\pm0.03}$ \\
         IDLike & $9.71^{\pm0.60}$ & $98.05^{\pm0.07}$ & $38.93^{\pm0.10}$ & $90.54^{\pm0.68}$ & $47.06^{\pm1.44}$ & $88.06^{\pm0.90}$ & $32.82^{\pm5.12}$ & $91.89^{\pm1.49}$ & $32.12^{\pm1.09}$ & $92.14^{\pm0.01}$\\
         NegPrompt & $37.79^{\pm0.11}$ & $90.49^{\pm0.01}$ & $32.11^{\pm3.77}$ & $92.25^{\pm1.00}$ & $35.52^{\pm0.41}$ & $91.16^{\pm0.03}$ & $43.93^{\pm9.09}$ & $88.38^{\pm3.31}$ & $37.34^{\pm1.41}$ & $90.57^{\pm0.59}$ \\
         LSN & $36.17^{\pm4.81}$ & $92.66^{\pm1.16}$ & $34.27^{\pm0.44}$ & $93.53^{\pm0.20}$ & $41.47^{\pm0.85}$ & $90.52^{\pm0.37}$ & $46.43^{\pm0.60}$ & $89.38^{\pm0.24}$ & $39.58^{\pm0.73}$ & $91.53^{\pm0.09}$ \\
         SCT & $13.94^{\pm0.68}$ & $95.86^{\pm0.28}$ & $20.55^{\pm1.07}$ & $95.33^{\pm0.12}$ & $29.86^{\pm0.67}$ & $92.24^{\pm0.05}$ & $41.51^{\pm0.48}$ & $89.06^{\pm0.09}$ & $\textbf{26.47}^{\pm0.39}$ & $\textbf{93.37}^{\pm0.07}$ \\

        \bottomrule[1.5pt]
    \end{tabular}}
    \label{tab:main_table1}
\end{table*}

\subsection{Main results}
\label{sec:main_results}
In this part, we present the major performance comparison with some representative baseline methods for OOD detection to demonstrate the effectiveness of the proposed SCT. Specifically, we consider several zero-shot methods as the performance reference based on the pretrained CLIP and some prompt tuning based methods for specific comparison on fine-tuning with few-shot ID data. Note that we leave the experiment results on 2-shot and 4-shot settings in Appendix~\ref{app:add_exp_result}.

\begin{table*}[t!]
    \caption{OOD detection performance on compatibility experiments. All methods are trained on the same backbone. $\uparrow$ indicates larger values are better, and $\downarrow$ indicates smaller values are better. Methods are run under multiple trials reporting the mean and standard deviation of the performance.} 
    \vspace{2mm}
    \centering
    \footnotesize
    \renewcommand\arraystretch{1.0}
    \resizebox{\textwidth}{!}{
    \begin{tabular}{lcccccccccc}
        \toprule[1.5pt]
         \multirow{3}*{\textbf{Method}} & \multicolumn{8}{c|}{\textbf{OOD Dataset}} \\
        %\cmidrule{3-8}
         ~ & \multicolumn{2}{c}{\textbf{iNaturalist}} & \multicolumn{2}{c}{\textbf{SUN}} & \multicolumn{2}{c}{\textbf{Places365}} &  \multicolumn{2}{c}{\textbf{Textures}} & \multicolumn{2}{c}{\textbf{Average}}\\
         ~ & FPR95$\downarrow$ & AUROC$\uparrow$ & FPR95$\downarrow$ & AUROC$\uparrow$ & FPR95$\downarrow$ & AUROC$\uparrow$  & FPR95$\downarrow$ & AUROC$\uparrow$  & FPR95$\downarrow$ & AUROC$\uparrow$ \\
        \midrule[0.6pt]
        ~ & \multicolumn{10}{c}{1-shot}\\
         LoCoOp & $28.81^{\pm2.78}$ & $94.05^{\pm0.72}$ & $25.76^{\pm0.53}$ & $94.51^{\pm0.19}$ & $33.68^{\pm0.62}$ & $91.59^{\pm0.23}$ & $51.53^{\pm0.42}$ & $86.85^{\pm0.24}$ & $34.94^{\pm0.65}$ & $91.75^{\pm0.23}$ \\
         SCT & $19.16^{\pm1.15}$ & $95.70^{\pm0.28}$ & $23.52^{\pm1.91}$ & $94.58^{\pm0.56}$ & $32.81^{\pm1.14}$ & $91.23^{\pm0.32}$ & $48.87^{\pm1.38}$ & $86.66^{\pm0.33}$ & $\textbf{31.09}^{\pm0.48}$ & $\textbf{92.04}^{\pm0.25}$ \\
         IDLike & $12.07^{\pm0.88}$ & $97.65^{\pm0.10}$ & $40.55^{\pm5.84}$ & $91.07^{\pm1.80}$ & $47.94^{\pm5.24}$ & $88.31^{\pm2.05}$ & $38.34^{\pm13.39}$ & $89.67^{\pm4.03}$ & $34.72^{\pm0.80}$ & $\textbf{91.67}^{\pm0.07}$ \\
         IDLike+SCT & $12.24^{\pm2.09}$ & $97.58^{\pm0.54}$ & $31.98^{\pm1.07}$ & $92.27^{\pm0.26}$ & $44.79^{\pm2.82}$ & $87.62^{\pm0.69}$ & $43.57^{\pm1.99}$ & $86.71^{\pm1.3}$ & $\textbf{33.14}^{\pm0.58}$ & $90.97^{\pm0.36}$\\
         LSN & $59.28^{\pm7.02}$ & $87.20^{\pm3.15}$ & $40.15^{\pm0.82}$ & $91.47^{\pm0.14}$ & $46.11^{\pm1.86}$ & $88.74^{\pm0.57}$ & $60.34^{\pm0.14}$ & $83.92^{\pm0.42}$ & $51.47^{\pm1.53}$ & $87.84^{\pm0.58}$ \\
         LSN+SCT & $51.38^{\pm1.24}$ & $89.54^{\pm0.14}$ & $35.60^{\pm0.53}$ & $92.48^{\pm0.09}$ & $40.84^{\pm0.59}$ & $90.29^{\pm0.10}$ & $55.24^{\pm0.01}$ & $86.00^{\pm0.03}$ & $\textbf{45.76}^{\pm0.59}$ & $\textbf{89.58}^{\pm0.09}$ \\
         \midrule[0.6pt]
         ~& \multicolumn{10}{c}{16-shot}\\
         LoCoOp & $17.58^{\pm2.22}$ & $96.30^{\pm0.57}$ & $22.82^{\pm0.34}$ & $95.20^{\pm0.06}$ & $32.21^{\pm0.53}$ & $92.03^{\pm0.20}$ & $45.27^{\pm0.95}$ & $88.86^{\pm0.26}$ & $29.47^{\pm0.29}$ & $93.10^{\pm0.03}$ \\
         SCT & $13.94^{\pm0.68}$ & $95.86^{\pm0.28}$ & $20.55^{\pm1.07}$ & $95.33^{\pm0.12}$ & $29.86^{\pm0.67}$ & $92.24^{\pm0.05}$ & $41.51^{\pm0.48}$ & $89.06^{\pm0.09}$ & $\textbf{26.47}^{\pm0.39}$ & $\textbf{93.37}^{\pm0.07}$ \\
         IDLike & $9.71^{\pm0.60}$ & $98.05^{\pm0.07}$ & $38.93^{\pm0.10}$ & $90.54^{\pm0.68}$ & $47.06^{\pm1.44}$ & $88.06^{\pm0.90}$ & $32.82^{\pm5.12}$ & $91.89^{\pm1.49}$ & $32.12^{\pm1.09}$ & $\textbf{92.14}^{\pm0.01}$\\
         IDLike+SCT & $3.41^{\pm1.88}$ & $98.97^{\pm0.38}$ & $33.98^{\pm2.30}$ & $89.24^{\pm2.18}$ & $37.81^{\pm0.44}$ & $87.49^{\pm1.33}$ & $29.49^{\pm2.58}$ & $90.07^{\pm0.35}$ & $\textbf{26.17}^{\pm0.65}$ & $91.44^{\pm0.69}$ \\

         LSN & $36.17^{\pm4.81}$ & $92.66^{\pm1.16}$ & $34.27^{\pm0.44}$ & $93.53^{\pm0.20}$ & $41.47^{\pm0.85}$ & $90.52^{\pm0.37}$ & $46.43^{\pm0.60}$ & $89.38^{\pm0.24}$ & $39.58^{\pm0.73}$ & $91.53^{\pm0.09}$ \\
         LSN+SCT & $35.14^{\pm0.89}$ & $92.64^{\pm0.06}$ & $29.80^{\pm1.65}$ & $94.06^{\pm0.30}$ & $37.00^{\pm0.90}$ & $91.25^{\pm0.17}$ & $45.45^{\pm1.34}$ & $89.46^{\pm0.46}$ & $\textbf{36.85}^{\pm0.08}$ & $\textbf{91.85}^{\pm0.01}$ \\

        \bottomrule[1.5pt]
    \end{tabular}}
    \label{tab:main_table2}
\end{table*}

\textbf{Comparisons on conventional OOD detection} In Table~\ref{tab:main_table1}, we present the overall results of the comparison between different baseline methods and SCT for OOD detection. Since the prompt tuning based methods engage the ID data during training, the model will generally gain better empirical performance on OOD detection, reflected by evaluation metrics like FPR95 and AUROC. IDLike, NegPrompt, and LSN all introduce a set of negative prompts for each ID class to learn negative semantics of ID objects using different strategies, which obtain different levels of detection performance gains. Without sacrificing much classification performance (i.e., ID classification accuracy) on ID data, as shown in Table~\ref{exp:id_accuracy}, our SCT can consistently achieve better OOD detection performance on the large-scale ImageNet-1k benchmark, which verifies the effectiveness of our methods with the newly proposed modulation factors. 

\textbf{Compatibility with other baselines.} In Table~\ref{tab:main_table2}, we report the results of compatibility experiments, in which we compare those prompt tuning based methods with their variants, incorporating our SCT to dynamically adjust the importance of OOD regularization from ID samples with different uncertainty levels. We can find that our SCT can consistently help them gain better or comparable OOD detection performance across two evaluation metrics while keeping the classification accuracy comparable with the vanilla prompt-tuned model, as shown in Appendix~\ref{app:add_exp_result}. 

\begin{table}[t]
  \begin{minipage}[h]{0.48\linewidth}
     
   \centering
    \captionof{table}{OOD detection performance comparison with LoCoOp on hard OOD detection tasks. Bold numbers represents superior results.}
    \label{exp:hard}
    \scriptsize
    \renewcommand\arraystretch{0.8}
    \resizebox{\linewidth}{!}{%
      \begin{tabular}{ccccc}
\toprule
\multicolumn{1}{l}{ID Dataset} & \multicolumn{1}{l}{OOD Dataset} & Method & \multicolumn{1}{c}{FPR95↓} & \multicolumn{1}{c}{AUROC↑} \\ 
\midrule
 & & LoCoOp & 28.20 & 92.75 \\
\multirow{-2}{*}{ImageNet-10} & \multirow{-2}{*}{ImageNet-20} & \cellcolor[HTML]{EFEFEF}SCT & \cellcolor[HTML]{EFEFEF}\textbf{25.10} & \cellcolor[HTML]{EFEFEF}\textbf{94.33} \\ 
\midrule
& & LoCoOp & 34.40 & 92.34 \\
\multirow{-2}{*}{ImageNet-20} & \multirow{-2}{*}{ImageNet-10} & \cellcolor[HTML]{EFEFEF}SCT & \cellcolor[HTML]{EFEFEF}\textbf{25.00} & \cellcolor[HTML]{EFEFEF}\textbf{94.95} \\ 
\midrule
& & LoCoOp & 30.08 & 93.00 \\
\multirow{-2}{*}{ImageNet-10} & \multirow{-2}{*}{ImageNet-100} & \cellcolor[HTML]{EFEFEF}SCT & \cellcolor[HTML]{EFEFEF}\textbf{26.64} & \cellcolor[HTML]{EFEFEF}\textbf{93.90} \\ 
\midrule
& & LoCoOp & 61.40 & 81.97 \\
\multirow{-2}{*}{ImageNet-100} & \multirow{-2}{*}{ImageNet-10} & \cellcolor[HTML]{EFEFEF}SCT & \cellcolor[HTML]{EFEFEF}\textbf{57.80} & \cellcolor[HTML]{EFEFEF}\textbf{82.60} \\ 
\bottomrule      \end{tabular}
    }
  \end{minipage}
  % \hspace{0.5cm} % Add some space between the table and the figure
  \hfill
  \begin{minipage}[h]{0.5\linewidth}
   \centering
    \captionof{table}{Ablation study on the modulation factors for the classification task and regularization term. Detailed results can be found in Appendix~\ref{app:add_exp_result}.}
    \label{tab:modulation_factors}
    %\resizebox{\linewidth}{!}{%
    \scriptsize
    \renewcommand\arraystretch{0.75}
      \begin{tabular}{cccccccc}
\toprule
\multicolumn{1}{c}{Shot} & \multicolumn{2}{c}{$\phi$} & \multicolumn{2}{c}{$\psi$} & \multicolumn{1}{c}{FPR95↓} & \multicolumn{1}{c}{AUROC↑} & ID-ACC↑ \\ 
\midrule
  & \multicolumn{2}{c}{\xmark} & \multicolumn{2}{c}{\xmark} & 34.94 & 91.75 & \textbf{69.03} \\
  & \multicolumn{2}{c}{\cmark} & \multicolumn{2}{c}{\xmark} & 31.14 & \textbf{92.35} & 68.60 \\
  & \multicolumn{2}{c}{\xmark} & \multicolumn{2}{c}{\cmark} & 31.90 & 91.74 & 68.70\\
  %\cmidrule{2-7}
  \multirow{-4}{*}{1-shot} & \multicolumn{2}{c}{\cmark} & \multicolumn{2}{c}{\cmark} & \textbf{31.09} & 92.04 & 68.80 \\
 \midrule
  & \multicolumn{2}{c}{\xmark} & \multicolumn{2}{c}{\xmark} & 29.47 & 93.10 & 71.43 \\
  & \multicolumn{2}{c}{\cmark} & \multicolumn{2}{c}{\xmark} & 29.30 & 92.66 & 71.50 \\
  & \multicolumn{2}{c}{\xmark} & \multicolumn{2}{c}{\cmark} & 28.94 & 92.62 & \textbf{71.90}\\
 %\cmidrule{2-7}
  \multirow{-4}{*}{16-shot} & \multicolumn{2}{c}{\cmark} & \multicolumn{2}{c}{\cmark} & \textbf{26.47} & \textbf{93.37} & 71.77\\
\bottomrule      \end{tabular}
    %}
  \end{minipage}

\end{table}

\textbf{Comparisons on hard OOD detection.}
Following the setup in MCM~\citep{ming2022delving}, we also explore the performance of SCT on hard OOD detection tasks, as shown in
Table~\ref{exp:hard}. SCT significantly outperforms LoCoOp in all four experimental settings, demonstrating that SCT has
strong discriminative power for semantically hard OOD data.

\subsection{Ablation study}
\label{sec:ablation}

In this part, we conduct various ablation experiments and further explorations to provide a thorough understanding of the characteristics of our proposed SCT. For the extra results and discussions (e.g., computational cost and social impact), we leave more details in Appendix~\ref{app:add_exp_result}.

\paragraph{Importance of the modulation factors in SCT.} 
As an important aspect of the learning objective of SCT in Eq.~\eqref{eq: SCT}, the newly introduced modulating factors in the learning objective of SCT conduct adaptive redirection of prompt tuning towards suitable tasks. Although each of modulating factors can seemingly achieve the redirection effect alone, we empirically find that both modulating factors play a significant role in enhancing OOD detection performance of VLMs, as shown in Table~\ref{tab:modulation_factors}.

\begin{figure}[t]
  \centering
  \begin{center}
    \subfigure[Regularization Weight]{
    \includegraphics[scale=0.138]{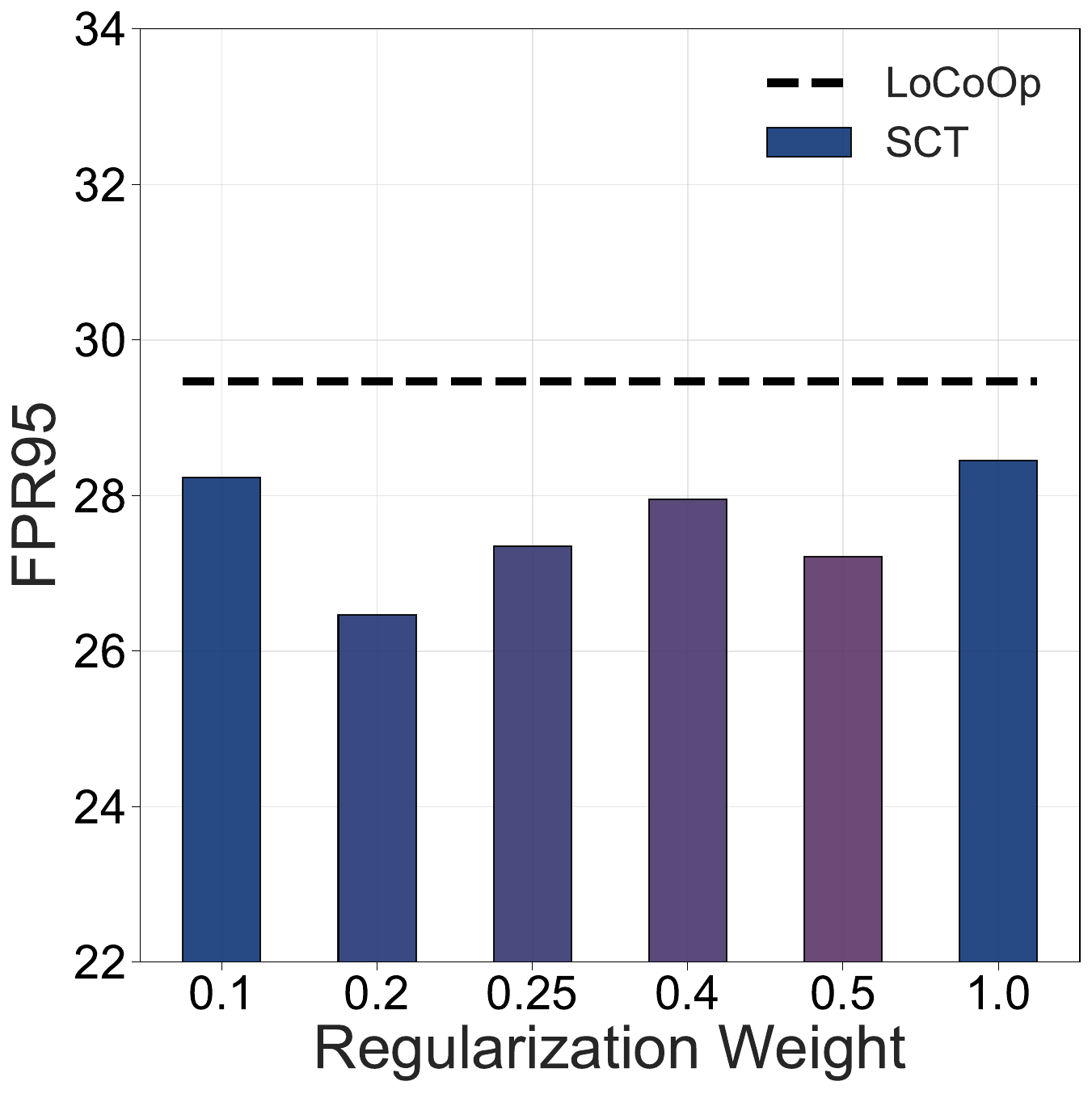}
    \label{fig4:a}
    }
    \subfigure[Regularization Func.]{
    \includegraphics[scale=0.138]{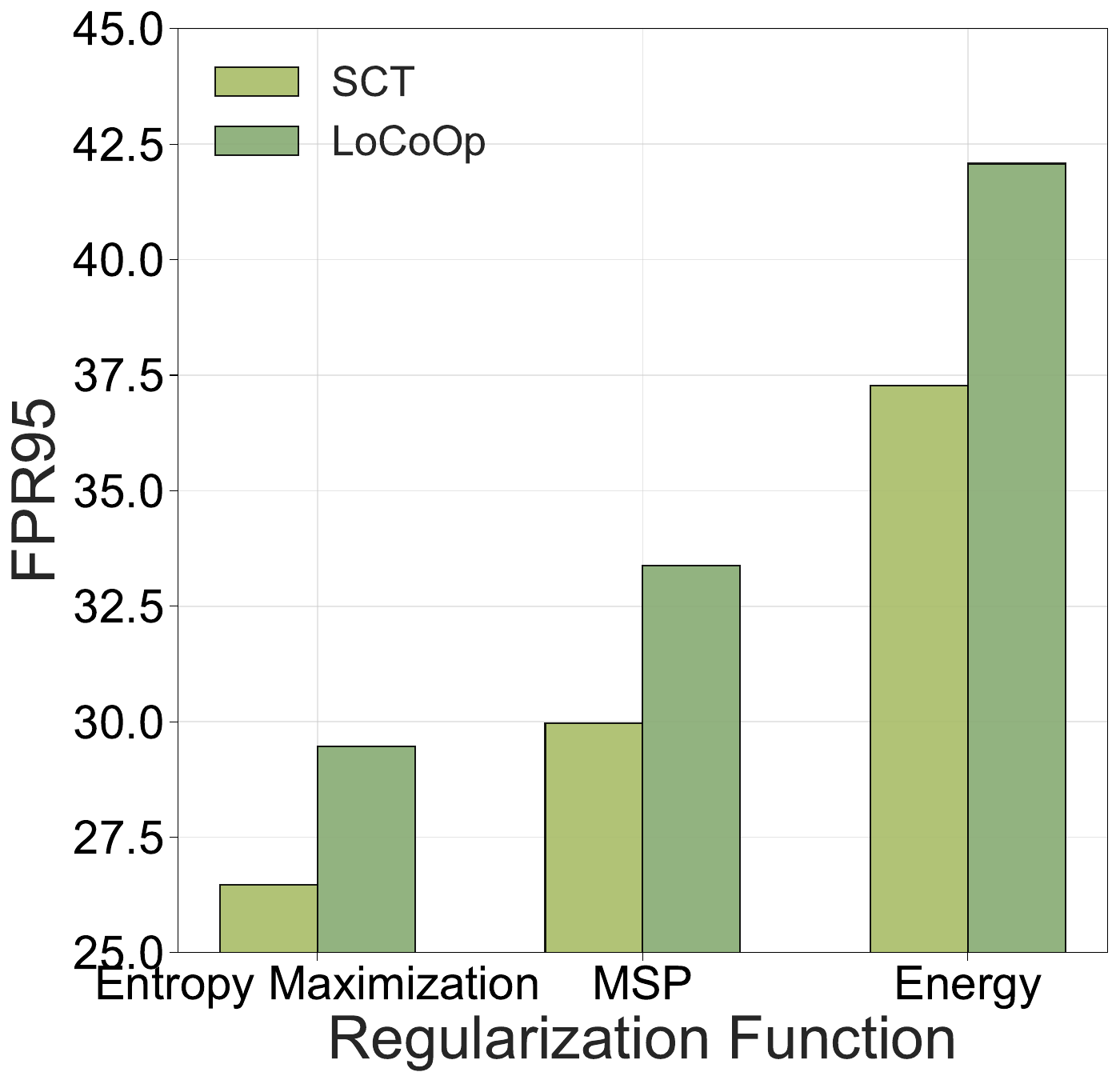}
    \label{fig4:b}
    }
    \subfigure[Architecture]{
    \includegraphics[scale=0.138]{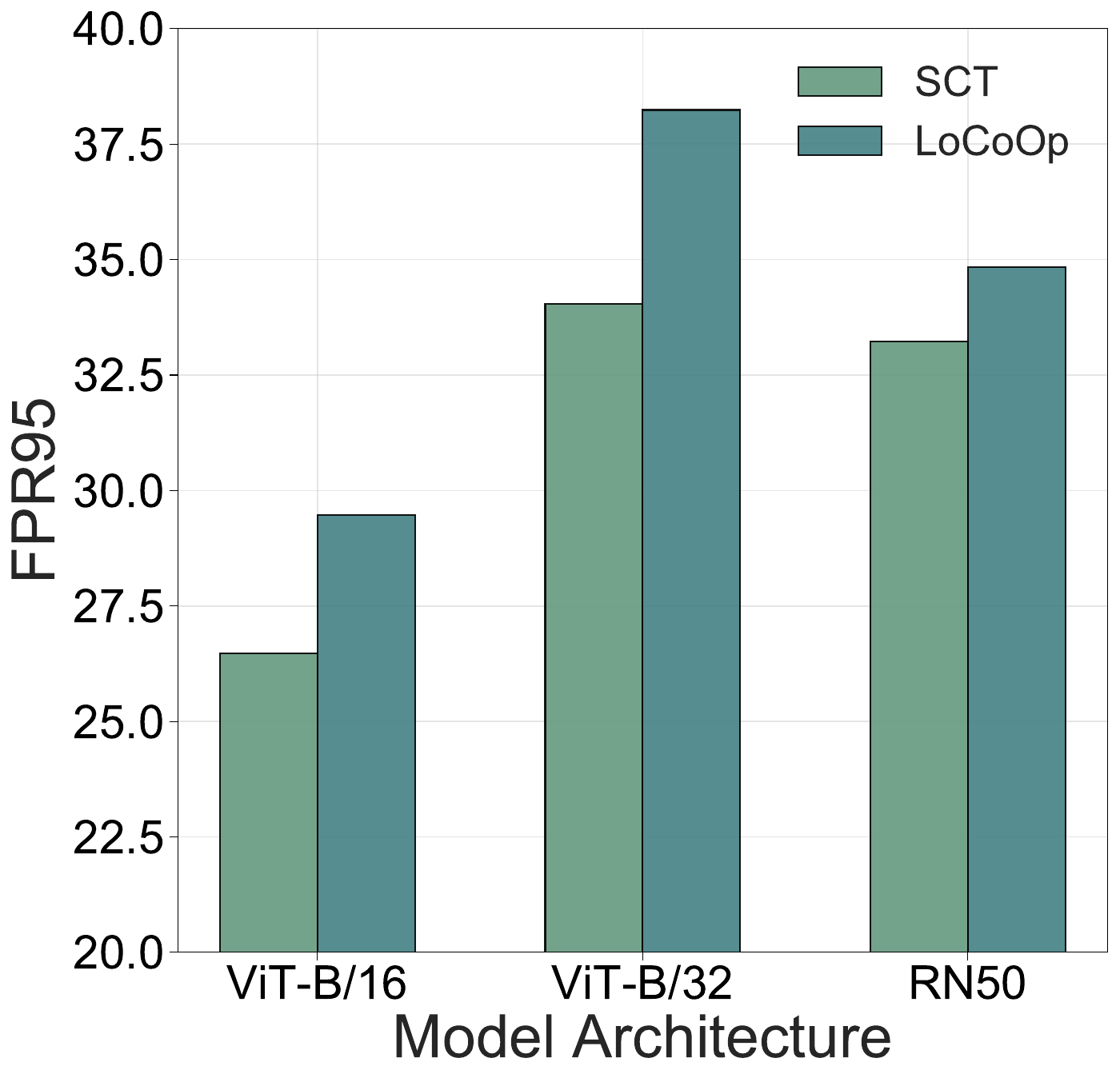}
    \label{fig4:c}
    }
    \subfigure[Extraction Method]{
    \includegraphics[scale=0.138]{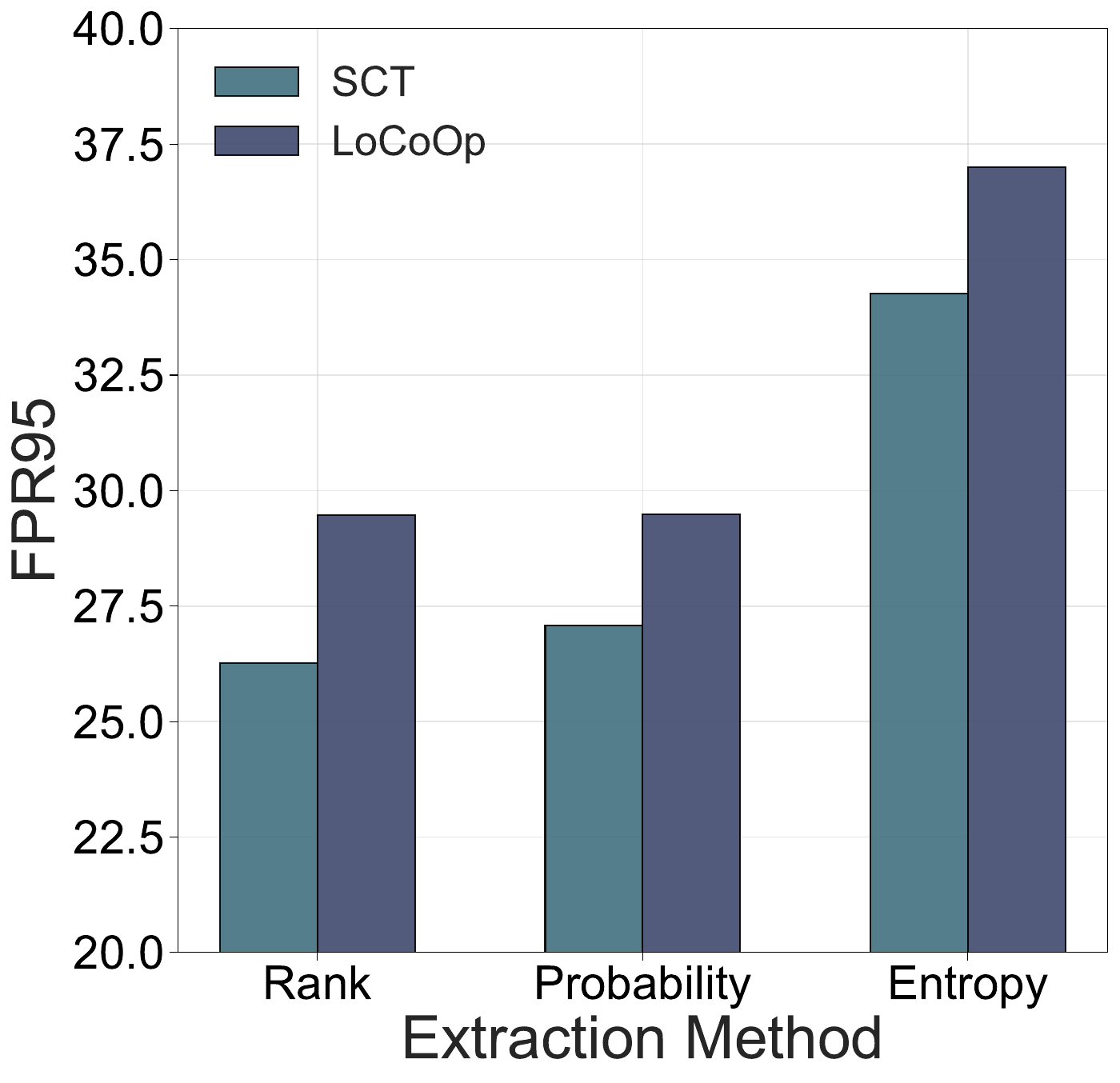}
    \label{fig4:d}
    }
    \end{center}
    %\vspace{-2mm}
  \caption{Ablation study. (a) performance of using different regularization weights $\lambda$; (b) exploration of different regularization functions for OOD regularization; (c) using different CLIP architectures; (d) comparison of different methods for extracting OOD features.}
  \label{fig4:ablation}
  \vspace{-0mm}
\end{figure}

\paragraph{Influence of regularization weight in OOD regularization.} 
The regularization weight $\lambda$ controls the contribution of OOD regularization to the prompt learning process like the role of the two modulation factors.  In Figure~\ref{fig4:a}, we show the performance by varying the ratio $\lambda=1$. It is worth noting that setting high regularization weights such as $\lambda=1$ may even degrade the performance, indicating that the ID classification task should be attached more importance for better OOD detection.

\paragraph{Generality of using different OOD regularization functions.} 
Since SCT introduces a general learning framework of prompt tuning for OOD detection, the specific realization for the OOD regularization function can have multiple choices (e.g., MSP~\citep{hendrycks2016baseline} or Energy score function~\citep{liu2020energy}).  Here we report the performance using different OOD regularization functions in Figure~\ref{fig4:b}, where they have different performance improvements compared with the original LoCoOp baseline. Specifically, the energy regularization needs tuning the two energy threshold hyperparameters $m_{in}$ and $m_{out}$, limiting its advantages over other regularization functions.

\paragraph{Implementation with different CLIP architectures.} 
We evaluate SCT with different VLM architectures and
the results are shown in Figure~\ref{fig4:c}. The result illustrates that the larger backbone boosts the performance of OOD detection and also shows SCT can outperform LoCoOp across various VLM architectures. It is important to note that we
take the same hyperparameters across all architectures, demonstrating the robustness of SCT
hyperparameters on different VLM architectures.

\paragraph{Comparison between different methods for extracting OOD features.} In Figure~\ref{fig4:d}, we perform the comparison of our method adopting different methods for extracting OOD features, including the probability-based and entropy-based methods. The results verify the superiority of SCT on OOD detection across all the OOD feature extraction methods. Specifically, the probability-based method and entropy-based method both have a threshold hyperparameter to discriminate between ID and OOD features. The sensitivity to hyperparameters of different methods might be the reason behind the different OOD detection performance. For the entropy-based method, the performance is significantly poor since it is challenging to determine the appropriate threshold~\citep{miyai2024locoop}.

\section{Discussions and limitations}
\label{discuss_limit}

\paragraph{Comparisons with advanced post-hoc methods.} Recently, advanced post-hoc approaches~\citep{djurisic2022extremely,xu2023scaling} exhibit comparable OOD detection performance to tuning based methods, despite the lack of training data. However, prompt tuning based methods can leverage the generalization ability of VLMs to better fit the domains of the downstream tasks given only few-shot ID training data. What's more, post-hoc methods and prompt tuning based methods are compatible with each other, further boosting the OOD detection performance. We provide more detailed discussions, including empirical experiments, on the compatibility of SCT with advanced post-hoc methods in Appendix~\ref{app:add_exp_result}. Furthermore, future research efforts into post-hoc calibration methods for prompt tuning based OOD detection could also contribute to the community.

\paragraph{Performance improvement on the AUROC metric.} The experiment results in Table~\ref{tab:main_table1} and ~\ref{tab:main_table2} indicate that the improvements of SCT over LoCoOp on AUROC are less notable than the FPR95 metric. However, as shown in the comparison of different baselines in Table~\ref{tab:main_table1}, the improvement space for FPR95 is significantly larger than AUROC. Therefore, the progress on these two metrics should not be treated equally. Nevertheless, further investigation is necessary to enhance the OOD detection performance specifically targeting improvement on the AUROC metric.

\paragraph{The sensitivity to training data under the few-shot setting.} VLMs are exposed to limited ID data samples under the few-shot scenarios, which means that the OOD detection performance of prompt tuning based methods, including SCT, can be susceptible to the quality of limited ID training data in practice. Exploration of selection strategies of suitable training data or overcoming the data sensitivity inherent in the few-shot setting could further facilitate the practical application of prompt tuning based OOD detection methods in real-world scenarios.

\paragraph{Theoretical analysis of the proposed method.} Although we propose a novel framework to mitigate the problem of invalid extracted OOD features, we have not yet provided sufficient theoretical analysis to prove the effectiveness of our method. We choose the linear function as the modulation function for the sake of simplicity, instead of based on theoretical justification. Conducting theoretical analyses on the relationship between sample uncertainty and OOD detection performance of prompt tuning based methods under few-shot settings is also a potential direction for future work.
.

\section{Conclusion}
In this paper, we propose a novel learning framework, i.e., \textit{Self-Calibrated Tuning (SCT)}, that improves the OOD detection capability of VLMs with only the given ID training data. To mitigate the problem caused by invalid OOD features mined from ID data, SCT introduces two modulating factors to the original learning objective to conduct adaptive redirection of prompt tuning process between the tasks of ID classification and OOD regularization. Through the redirection effect, our method calibrates the impact of OOD features extracted from different ID samples based on the sample uncertainty estimation during the training process, which facilitates the model learning from imperfect surrogate OOD features for OOD regularization. We have conducted extensive experiments to demonstrate the effectiveness of SCT and its compatibility with a range of prompt tuning based methods, along with various ablation studies and further explorations to characterize the framework.

\begin{ack}
GY and JCY were supported by the National Key R\&D Program of China (No. 2022ZD0160703), 111 plan (No. BP0719010) and National Natural Science Foundation of China (No. 62306178). JNZ and BH were supported by NSFC General Program No. 62376235, Guangdong Basic and Applied Basic Research Foundation Nos. 2022A1515011652 and 2024A1515012399, RIKEN Collaborative Research Fund, HKBU Faculty Niche Research Areas No. RC-FNRA-IG/22-23/SCI/04, and HKBU CSD Departmental Incentive Scheme.

% Use unnumbered first level headings for the acknowledgments. All acknowledgments
% More information about this disclosure can be found at: \url{https://neurips.cc/Conferences/2024/PaperInformation/FundingDisclosure}.

% Do {\bf not} include this section in the anonymized submission, only in the final paper. You can use the \texttt{ack} environment provided in the style file to automatically hide this section in the anonymized submission.
\end{ack}

\clearpage
\bibliography{neurips_2024}
\bibliographystyle{plainnat}

\clearpage

%%%%%%%%%%%%%%%%%%%%%%%%%%%%%%%%%%%%%%%%%%%%%%%%%%%%%%%%%%%%
\appendix

\section{Appendix / Supplemental Material}
\label{app:begin}

The whole Appendix is organized as follows. In Appendix~\ref{app:baseline_info}, we present the detailed definitions and implementation of zero-shot, post-hoc methods and several prompt tuning based methods that are considered in our experiments. In Appendix~\ref{app:exp_allapp}, we provide our extra experimental details and more comprehensive results with further discussion on the underlying implications. 
%In Appendix~\ref{app:related_compare}, we conduct a detailed comparison with related work. 
Finally, in Appendix~\ref{app:impact}, we discuss the potential broader impact and limitations of our work.

% Optionally include supplemental material (complete proofs, additional experiments and plots) in appendix.
% All such materials \textbf{SHOULD be included in the main submission.}

\section*{Reproducibility statement}

To ensure reproducibility, we outline several key aspects about the experiments below:

\begin{itemize}
    \item \textbf{Datasets.}  The datasets we used are all publicly accessible, which are introduced in Section~\ref{sec:setup}. 
    \item \textbf{Open source.} The source code is publicly available at \url{https://github.com/tmlr-group/SCT}. We provide a backbone for our experiments as well as several auxiliary components, such as OOD detection performance evaluation.
    \item \textbf{Environment.} All experiments are conducted with multiple runs on NVIDIA GeForce RTX 3090 GPUs with Python 3.8 and PyTorch 1.12.
\end{itemize}

\subsection{Details about considered baselines}
\label{app:baseline_info}

In this section, we present the details about the baselines, including zero-shot, post-hoc methods, and several prompt tuning based methods, as well as related hyper-parameters that are considered in our work.

\paragraph{MSP.} \citet{hendrycks2016baseline} proposes to utilize maximum softmax probability(MSP) as the scoring function to differentiate between ID and OOD samples, of which the definition is as follows,
\begin{equation}
    S_\text{MSP}(\boldsymbol{x}; f) = \max_m P(y = m | \boldsymbol{x}; f) = \max~\texttt{softmax}(f(\boldsymbol{x}))
\end{equation}
where $f$ represents a given well-trained model and $m$ is one of the classes $\mathcal{Y}=\{1,\ldots, M\}$. A higher MSP score signifies a greater probability that a sample belongs to the in-distribution distribution, indicating the model's confidence on the sample. 

\paragraph{ODIN.} \citet{liang2017enhancing} designs the ODIN score function, utilizing the temperature scaling and minor perturbations to test samples to widen the gap between the distributions of ID and OOD data. The ODIN score is defined as follows,
\begin{equation}
    S_\text{ODIN}(\boldsymbol{x}; f) = \max_m P(y = m | \boldsymbol{\tilde{x}}; f) = \max~\texttt{softmax}(\frac{f(\boldsymbol{\tilde{x}})}{T})
\end{equation}
where $\tilde{x}$ denotes the perturbed samples (controled by $\epsilon$) and $T$ denotes the temperature.

\paragraph{Energy.} \citet{liu2020energy} proposes to use the Energy of the prediction logits to discriminate between the ID and OOD samples. The Energy score is formulated as follows,
\begin{equation}
    S_\text{Energy}(\boldsymbol{x}; f) = -T  \log \sum \limits_{m = 1}^M e^{f(\boldsymbol{x})_c / T}
\end{equation}
where $T$ denotes the temperature parameter. As theoretically proved in \citet{liu2020energy}, a lower Energy score represents a higher probability for a sample to belong to ID. 

\paragraph{ReAct.} ~\citet{sun2021react} designs a simple and effective approaches, named Rectified
Activations (ReAct), to alleviate model overconfidence on out-of-distribution data. This work observes that OOD samples can induce unusually high unit activation in the deep layer of neural networks. ReAct improves OOD detection by simply rectifying the activations at an upper limit $c>0$, which can be performed on a pretrained model without any modification to training.

\paragraph{MaxLogit.} ~\citet{hendrycks2019scaling} demonstrates that the conventional MSP OOD detector does not scale well to challenging large-scale multiclass and multi-label OOD detection setting and proposes a surprisingly simple detector based on the maximum logit, called MaxLogit, that greatly outperforms previous post-hoc methods. The definition of MaxLogit is as follows,
\begin{equation}
    S_\text{MaxLogit}(\boldsymbol{x}; f) = -\max_m f(\boldsymbol{x})_m
\end{equation}

\paragraph{MCM.} ~\citet{ming2022delving} explores the potential of large-scale vision-language models like CLIP for OOD detection and proposes a simple yet effective zero-shot OOD detection method called Maximum Concept Matching (MCM), which is based on aligning visual features with textual concepts. This method characterizes OOD uncertainty by appropriately scaling of the distance from the visual feature to the nearest ID prototype, of which the formulation is as follows, 
\begin{equation}
S_{\mathrm{MCM}} = \max _{m} \frac{\exp \left(\operatorname{sim}\left(\boldsymbol{f}, \boldsymbol{g}_m\right) / \tau\right)}{\sum_{m^{\prime}=1}^M \exp \left(\operatorname{sim}\left(\boldsymbol{f}, \boldsymbol{g}_{m^{\prime}}\right) / \tau\right)}. 
\end{equation}

where $\tau$ denotes the temperature parameter, $\boldsymbol{f}$ and $\boldsymbol{g}_m$ represents image and text features respectively, and $sim()$ represents the cosine similarity between the image and text features.

\paragraph{GL-MCM.} ~\citet{miyai2023zero} designs a zero-shot OOD detection method called Global-Local Maximum Concept Matching(GL-MCM) that leverages the maximum softmax score of both global and
local image features and designs. Utilizing the matching score with local
features can compensate for the low global matching score and produce more accurate ID confidence. The detailed formulation of GL-MCM is shown in Eq~\eqref{eq:gl_mcm}.

\paragraph{CoOp.}
Drawing inspiration from the success in prompt learning within natural language processing, ~\citet{zhou2021learning} proposes Context Optimization (CoOp), a simple framework specifically to
adapt vision-language models like CLIP for downstream image recognition. Concretely, CoOp represents the context words of a prompt as learnable vectors while keeping the entire pretrained encoders fixed. The specific framework of CoOp is presented in Section~\ref{sec:prelimi}.

\paragraph{LoCoOp.} ~\citet{miyai2024locoop} introduces an approach named Local regularized Context Optimization (LoCoOp) to enhance the OOD detection performance of prompt tuning. It first extracts ID-irreverent contexts from CLIP’s local features and utilizes it as OOD features to perform OOD regularization with entropy maximization. The details of this method are provided in Section~\ref{sec:prelimi}

\paragraph{IDLike.} ~\citet{bai2023id} designs a new framework of prompt tuning for OOD detection to focus on challenging OOD detection scenarios. It first constructs surrogate outliers from ID samples by conducting multiple random cropping on ID data and filtering them based on their cosine similarity with textual features of ID classes. Then it introduces a set of OOD prompts along with conventional ID prompts and optimizes them with a new loss function that consists of in-distribution loss, out-of-distribution loss and diversification regularization.

\paragraph{NegPrompt.} ~\citet{li2024learning} proposes a prompt tuning based OOD detection method, named NegPrompt, which learns a set of negative prompts for each ID class, with only ID data, to capture the negative semantics associated with these classes. The training process is divided into two stages. In the first stage, the model learns the positive prompts. In the second stage, the positive prompts are kept frozen and the negative prompts are optimized via three loss functions that enforce the separation between negative prompts and ID images, a proper degree of similarity between negative and positive prompts, and the diversity of the negative prompts.

\paragraph{LSN.} ~\citet{nie2023out} reveals that CLIP cannot fully understand the negative semantics of textual information and proposes to learn a set of negative prompts for each class to alleviate this problem. The learned positive prompt (for all classes) and negative prompts (for each class) are utilized simultaneously to calculate similarity and dissimilarity in the feature space, thereby enhancing the accuracy of OOD sample detection.

\subsection{In-depth comparison between SCT and hard example mining.}
The part of ID classification in the learning framework of SCT bears a strong resemblance to hard example mining methods, such as focal loss~\citep{lin2017focal}. In this section, we clarify the novelty and insights of our SCT by analyzing the difference between SCT and hard example mining as follows.

Conceptually, the motivation of SCT is to mitigate the problem of unreliable OOD features in prompt tuning based OOD detection methods. Generally, these methods rely on the ID-irrelevant local context extracted by VLMs as the surrogate OOD features to perform regularization, the quality of which is greatly affected by the inaccurate foreground-background decomposition of VLMs. As shown in Figure~\ref{fig1:illustration} and Figure~\ref{fig:problem}, although VLMs can mask out some ID-related regions (shown as the grey patches of images), large portions of the extracted OOD features (shown as the colored patches of images) obviously belongs to ID features.

Empirically, we find that the quality of extracted OOD features significantly correlated with the uncertainty level of ID data. As illustrated in the left panel of Figure~\ref{fig2:emprical_demo}, the extracted OOD features become more inaccurate as the uncertainty increases. In the right panel of Figure~\ref{fig2:emprical_demo}, we train LoCoOp on multiple data groups with different uncertainty levels. The results demonstrate that the OOD detection performance of LoCoOp can be significantly impacted by the uncertainty level of ID data. Therefore, to mitigate the issue of unreliable OOD features, we propose SCT to calibrate the influence of OOD regularization from different ID samples based on their uncertainty level.

Technically, despite the simple design, SCT is significantly different from hard sample mining. The latter conducts reweighting directly on the samples based on the classification difficulty during training. The former adaptively adjusts the importance between the two components of the original learning objectives for every single sample. Data with high uncertainty are directly down-weighted in hard sample mining while they are utilized more for OOD regularization in SCT. As shown in Table~\ref{tab:modulation_factors}, under 16-shot ID data, the OOD detection performance of simply assigning $1-p(y|x)$ to $L_{ce}$ (denoted as $\phi$ $\checkmark$ and $\psi$ $\times$) are significantly inferior to SCT (denoted as $\phi$ $\checkmark$ and $\psi$ $\checkmark$), demonstrating the difference of SCT and hard sample mining.

\subsection{Additional experimental results and further discussion}
\label{app:exp_allapp}

In this section, we provide more experiment results from various perspectives to characterize our proposed SCT. First, we introduce the additional experimental setups for the empirical verification in previous figures and our learning framework. Second, we offer more detailed results and analyses of our method in comparison to other advanced baselines. Finally, more demonstrations of the motivation of our method are provided.

\subsubsection{Additional experimental setups} 
\label{app:add_exp_setup}

\paragraph{Figure~\ref{fig2:emprical_demo}.} In the right panel of Figure~\ref{fig2:emprical_demo}, we conduct experiments to investigate the relationship between sample uncertainty and OOD detection performance of prompt tuning based methods. We first calculate the prediction probability for ground-truth labels of all the training samples in a 64-shot training set using a prompt-tuned CLIP model. This model is prompt-tuned with LoCoOp on a 4-shot training set which contains no overlapping samples with the 64-shot set. We use the prediction probability for ground-truth labels to represent uncertainty. High-uncertainty samples are assigned low prediction probability for their ground-truth labels and vice versa. We choose the data with the lowest and highest uncertainty for every ID class to generate two data groups of specific shots with different uncertainty levels respectively, and train the model with LoCoOp on these two data groups.

\paragraph{Figure~\ref{fig4:ablation}.} In Figure~\ref{fig4:b}, we explore different regularization functions to perform OOD regularization, including entropy maximization~\citep{miyai2024locoop}, cross-entropy loss to the uniform distribution~\citep{hendrycks2018deep,ming2022poem} and the energy-based function~\citep{liu2020energy}, under the 16-shot setting. In Figure~\ref{fig4:a}, Figure~\ref{fig4:c} and Figure~\ref{fig4:d}, we train the models of all the architectures with 16-shot training datasets. In Figure~\ref{fig4:d}, we evaluate the performance of SCT with various methods for ID-irrelevant region extraction. To be specific, we consider three different methods, including the ranking-based method,  probability-based method, and entropy-based method. Following the setups in ~\citep{miyai2024locoop}, for the entropy-based method, we extract local regions where the entropy of $\boldsymbol{p}{_i}(\boldsymbol{x})$ is lower than $\frac{\log M}{2}$ since it is the half value of the maximum entropy of $M$-dimensional probabilities, as done in previous studies~\citep{saito2020universal}. For the probability-based method, we extract regions where  $p_i(y|\boldsymbol{x})$ is lower than $1/M$ simply because the ID dataset has $M$ classes.

\subsubsection{Additional experimental results and illustrations} 
\label{app:add_exp_result}

\paragraph{Experiments on multiple few-shot settings.} In order to further understand the effectiveness of our SCT on different few-shot settings, we report the evaluation results of our SCT with 2-shot and 4-shot training datasets, which are summarized in Table~\ref{exp:fewshot}. The results demonstrate that SCT can gain significant improvement on OOD detection under all the few-shot settings.

\begin{table*}[t!]
    \caption{OOD detection performance of SCT under various few-shot settings. All methods are trained on the same backbone CLIP-ViT-B/16. Bold numbers are superior results. $\uparrow$ indicates larger values are better, and $\downarrow$ indicates smaller values are better.} 
    \vspace{2mm}
    \centering
    \renewcommand\arraystretch{1.0}
    \resizebox{\textwidth}{!}{
    \begin{tabular}{lllllllllll}
        \toprule[1.5pt]
         \multirow{3}*{\textbf{Method}} & \multicolumn{8}{c|}{\textbf{OOD dataset}}\\
        %\cmidrule{3-8}
         ~ & \multicolumn{2}{c}{\textbf{iNaturalist}} & \multicolumn{2}{c}{\textbf{SUN}} & \multicolumn{2}{c}{\textbf{Places365}} &  \multicolumn{2}{c}{\textbf{Textures}} & \multicolumn{2}{c}{\textbf{Average}}\\
         ~ & FPR95$\downarrow$ & AUROC$\uparrow$ & FPR95$\downarrow$ & AUROC$\uparrow$ & FPR95$\downarrow$ & AUROC$\uparrow$  & FPR95$\downarrow$ & AUROC$\uparrow$  & FPR95$\downarrow$ & AUROC$\uparrow$\\
        \midrule[0.6pt]
        ~ & \multicolumn{10}{c}{2-shot}\\
         LoCoOp & 19.45 & 96.01 & 23.02 & 95.08 & 32.32 & 91.96 & 46.35 & 88.60 & 30.28 & 92.91 \\
         SCT & 16.99 & 96.15 & 21.68 & 94.79 & 31.01 & 92.21 & 42.62 & 88.64 & 28.08 & 92.95 \\
         \midrule[0.6pt]
        ~& \multicolumn{10}{c}{4-shot}\\
         LoCoOp & 17.32 & 96.26 & 23.78 & 94.82 & 32.19 & 91.85 & 44.41 & 89.22 & 29.42 & 93.04 \\
        SCT & 13.88 & 97.03 & 22.13 & 94.85 & 30.20 & 92.09 & 45.53 & 87.96 & 27.93 & 92.98 \\
        \bottomrule[1.5pt]
    \end{tabular}}
    \label{exp:fewshot}
\end{table*}

\paragraph{ID classification performance of baselines and SCT in Table~\ref{tab:main_table1} and Table~\ref{tab:main_table2}.} To evaluate the performance of all the considered baselines and SCT on the original classification task, we report the classification accuracy on the ID test set in Table~\ref{exp:id_accuracy}. Since NegPrompt and LSN learn positive prompts and negative prompts separately, they have the same ID classification performance as vanilla CoOp. The results show that SCT maintains comparable ID classification performance compared with vanilla prompt tuning CoOp and other advanced prompt tuning based OOD detection methods while achieving more effective OOD detection.

\begin{table*}[t!]
    \caption{ID classification performance of SCT and all the considered baselines. All methods are trained on the same backbone CLIP-ViT-B/16.} 
    \vspace{2mm}
    \centering
    \renewcommand\arraystretch{1.0}
    \resizebox{0.5\textwidth}{!}{
    \begin{tabular}{lll}
        \toprule[1.5pt]
         \multirow{2}*{\textbf{Method}} & \multicolumn{2}{c}{\textbf{ID Accuracy}}\\
        %\cmidrule{3-8}
         ~ & \multicolumn{1}{c}{1-shot} & \multicolumn{1}{c}{16-shot} \\

        \midrule[0.6pt]
        \textit{Zero-shot methods} \\
         MCM & \multicolumn{2}{c}{66.7} \\
         GL-MCM & \multicolumn{2}{c}{66.7} \\
         \midrule[0.6pt]
         \textit{CLIP-based post-hoc methods} \\
         MSP & \multicolumn{2}{c}{66.7} \\
         ODIN & \multicolumn{2}{c}{66.7} \\
         Energy & \multicolumn{2}{c}{66.7} \\
         ReAct & \multicolumn{2}{c}{66.7} \\
         MaxLogit & \multicolumn{2}{c}{66.7} \\
         \midrule[0.6pt]
         \textit{Prompt tuning based methods} \\
         CoOp & 68.83 & 71.93 \\
         LoCoOp & 69.03 & 71.43 \\
         IDLike & 68.90 & 71.04 \\
         NegPrompt & 68.83 & 71.93 \\
         LSN & 68.83 & 71.93 \\
         SCT & 68.80 & 71.77  \\
         IDLike+SCT & 69.70 & 71.14\\
         LSN+SCT & 68.98 & 71.50 \\
        \bottomrule[1.5pt]
    \end{tabular}}
    \label{exp:id_accuracy}
\end{table*}

\paragraph{Fine-grained results of OOD test dataset for experiments in Table~\ref{tab:modulation_factors}.} To further evaluate the influence of both modulating factors introduced in Eq~\eqref{eq: SCT} on the performance on every OOD test test, we present the fine-grained OOD detection results in Table~\ref{exp:fine_grain_modulation}. We observe from the results that although each of the modulation factors can individually contribute to redirect model training, their combination maximizes the improvement on OOD detection, which achieves the best results on most OOD test sets and shot numbers.

\begin{table*}[t!]
    \caption{Fine-grained results of OOD detection performance for experiments in Table~\ref{tab:modulation_factors}. All methods are trained on the same backbone CLIP-ViT-B/16. Bold numbers are superior results. $\uparrow$ indicates larger values are better, and $\downarrow$ indicates smaller values are better.} 
    \vspace{2mm}
    \centering
    \renewcommand\arraystretch{1.0}
    \resizebox{\textwidth}{!}{
    \begin{tabular}{llllllllllll}
        \toprule[1.5pt]
         \multirow{3}*{\textbf{$\phi$}} & \multirow{3}*{\textbf{$\psi$}} & \multicolumn{8}{c|}{\textbf{OOD dataset}}\\
        %\cmidrule{3-8}
         ~ & & \multicolumn{2}{c}{\textbf{iNaturalist}} & \multicolumn{2}{c}{\textbf{SUN}} & \multicolumn{2}{c}{\textbf{Places365}} &  \multicolumn{2}{c}{\textbf{Textures}} & \multicolumn{2}{c}{\textbf{Average}}\\
         ~ & & FPR95$\downarrow$ & AUROC$\uparrow$ & FPR95$\downarrow$ & AUROC$\uparrow$ & FPR95$\downarrow$ & AUROC$\uparrow$  & FPR95$\downarrow$ & AUROC$\uparrow$  & FPR95$\downarrow$ & AUROC$\uparrow$\\
        \midrule[0.6pt]
        ~ & & \multicolumn{10}{c}{1-shot}\\
        \xmark & \xmark & $28.81$ & $94.05$ & $25.76$ & $94.51$ & $33.68$ & $91.59$ & $51.53$ & $86.85$ & $34.94$ & $91.75$ \\
        \cmark & \xmark & 21.94 & 95.14 & 22.46 & 95.08 & 30.54 & 92.04 & 49.61 & 87.14 & 31.14 & \textbf{92.35} \\
        \xmark & \cmark & 19.26 & 95.72 & 23.70 & 94.44 & 33.56 & 91.04 & 51.10 & 85.77 & 31.90 & 91.74 \\
         \cmark & \cmark & $19.16$ & $95.70$ & $23.52$ & $94.58$ & $32.81$ & $91.23$ & $48.87$ & $86.66$ & $\textbf{31.09}$ & $92.04$ \\
         \midrule[0.6pt]
        ~ & & \multicolumn{10}{c}{16-shot}\\
        \xmark & \xmark  & $17.58$ & $96.30$ & $22.82$ & $95.20$ & $32.21$ & $92.03$ & $45.27$ & $88.86$ & $29.47$ & $93.10$ \\
         \cmark & \xmark & 18.14 & 94.50 & 22.42 & 94.04 & 31.90 & 91.18 & 44.72 & 90.09 & 29.30 & 92.66 \\
        \xmark & \cmark & 15.08 & 96.92 & 21.42 & 95.16 & 30.60 & 92.07 & 48.64 & 86.35 & 28.94 & 92.62 \\
       \cmark & \cmark & $13.94$ & $95.86$ & $20.55$ & $95.33$ & $29.86$ & $92.24$ & $41.51$ & $89.06$ & $\textbf{26.47}$ & $\textbf{93.37}$ \\
        \bottomrule[1.5pt]
    \end{tabular}}
    \label{exp:fine_grain_modulation}
\end{table*}

\paragraph{Experiments on other instantiations of modulation functions.} To explore the generality of our proposed learning framework, we consider other instantiations of modulation functions. Specifically, we consider the power, logarithmic, and trigonometric functions, which are formulated in Table~\ref{tab:modulation_function}. For the experiment on the power function, we set $\lambda=0.25$, $\alpha=0.5$ for 1-shot and $\lambda=0.25$, $\alpha=4$ for 16-shot. The experiment results, as shown in Table~\ref{tab:exp_instantiation}, indicate that all the considered choices of modulation function can achieve significantly better OOD detection performance than LoCoOp, which further empirically proves the effectiveness of the learning framework of SCT.

\begin{table*}[t!]
    \caption{Different implementations of modulation functions.} 
    \vspace{2mm}
    \centering
    \renewcommand\arraystretch{1.0}
    \resizebox{0.7\textwidth}{!}{
    \begin{tabular}{lcc}
        \toprule[1.5pt]
        Modulation function & $\phi$ & $\psi$ \\
        \midrule[0.6pt]
        \multirow{2}{*}{power} & \multirow{2}{*}{$(1-
        p(y|\boldsymbol{x};\boldsymbol{\omega}))^\alpha $} & \multirow{2}{*}{$ p(y|\boldsymbol{x};\boldsymbol{\omega})^\alpha $} \\ \\
        \multirow{2}{*}{logarithmic} & \multirow{2}{*}{ $ 1-\frac{\log{(p(y|\boldsymbol{x};\boldsymbol{\omega})+1})}{\log{2}} $ } & \multirow{2}{*}{ $ \frac{\log{(p(y|\boldsymbol{x};\boldsymbol{\omega})+1})}{\log{2}} $ }\\ \\
        \multirow{2}{*}{trigonometric} & \multirow{2}{*}{ $ \cos{(\frac{\pi}{2} p(y|\boldsymbol{x};\boldsymbol{\omega}))} $ } & \multirow{2}{*}{ $ \sin{(\frac{\pi}{2} p(y|\boldsymbol{x};\boldsymbol{\omega}))} $ } \\ \\

        \bottomrule[1.5pt]
    \end{tabular}}
    \label{tab:modulation_function}
\end{table*}

\begin{figure}[t]
  \centering
  \includegraphics[width=0.3\columnwidth]{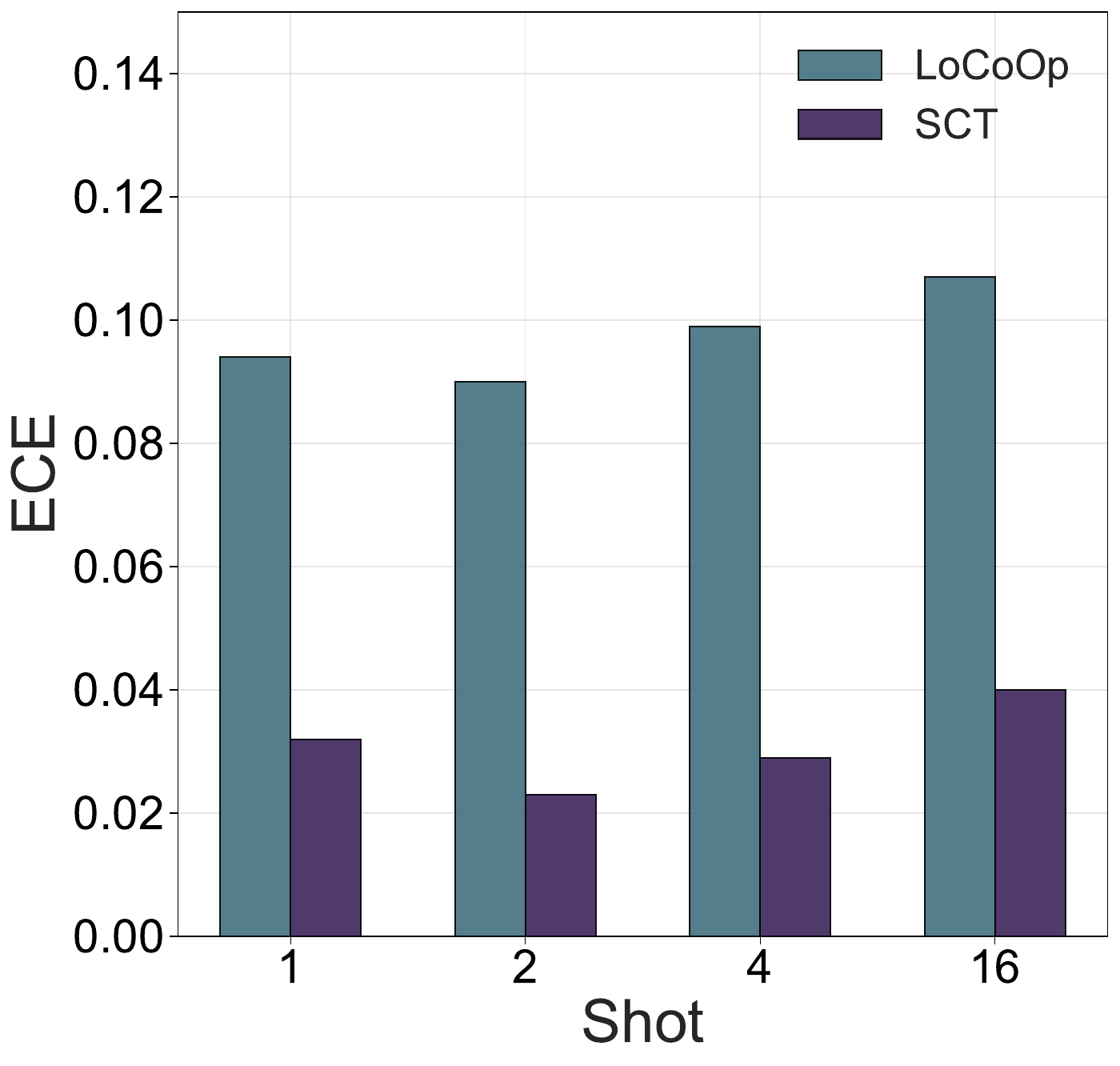}
  \caption{The comparison of calibration measured by ECE between SCT and LoCoOp trained with 1, 2, 4, 16 shots data. The evaluation is performed on the original validation set of ImageNet-1k.}
  \label{fig:ece}
\end{figure}

\begin{figure}[t]
  \centering
  \includegraphics[width=\columnwidth]{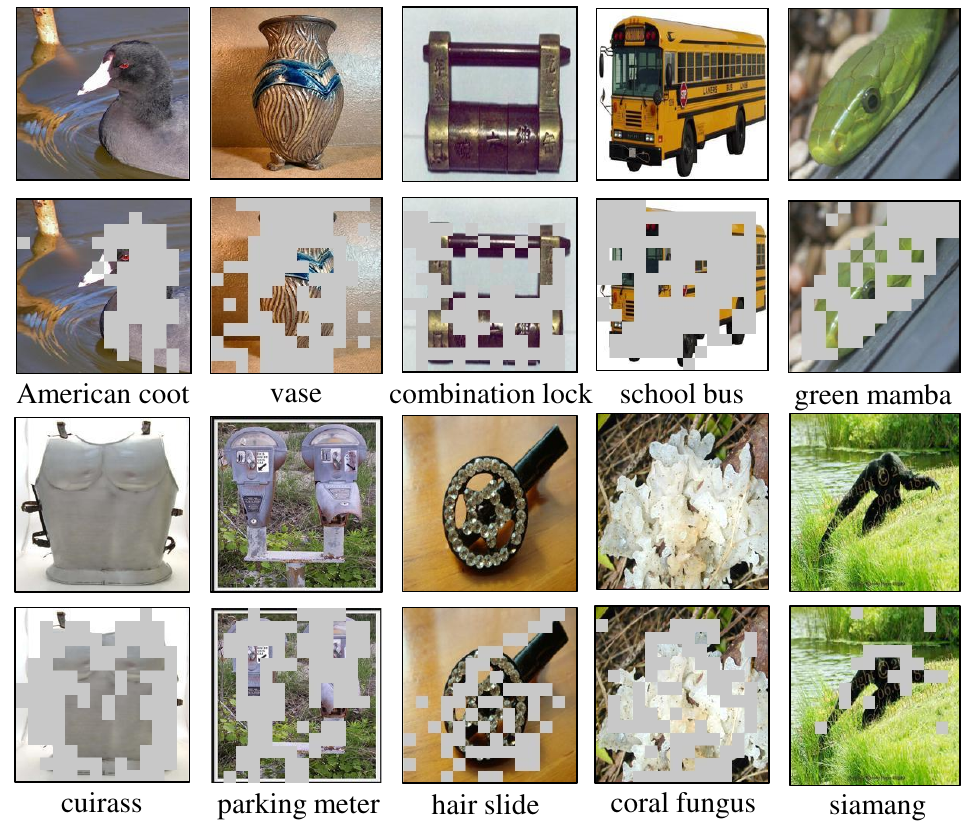}
  \caption{Examples of the invalid OOD features extracted by CLIP. The odd-numbered rows show the original images from ImageNet-1k and the even-numbered rows show the extracted ID-irrelevant context from the corresponding images. The ground-truth labels are annotated below every even-numbered row. Although CLIP can mask out some ID-related regions (shown as the gray patches of images), large portions of the extracted OOD features (shown as the colored patches of images) obviously belong to ID features.}
  \label{fig:problem}
\end{figure}

\begin{table*}[t!]
    \caption{OOD detection performance of different instantiations of modulation functions. All methods are trained on the same backbone CLIP-ViT-B/16. Bold numbers are superior results. $\uparrow$ indicates larger values are better, and $\downarrow$ indicates smaller values are better. SCT-L denotes SCT with the linear function, SCT-Pow denotes SCT with the power function, SCT-Log denotes SCT with the logarithmic function and SCT-Tri denotes SCT with the trigonometric function as the modulation function.} 
    \vspace{2mm}
    \centering
    \renewcommand\arraystretch{1.0}
    \resizebox{\textwidth}{!}{
    \begin{tabular}{lllllllllll}
        \toprule[1.5pt]
         \multirow{3}*{\textbf{Method}} & \multicolumn{8}{c|}{\textbf{OOD dataset}}\\
        %\cmidrule{3-8}
         ~ & \multicolumn{2}{c}{\textbf{iNaturalist}} & \multicolumn{2}{c}{\textbf{SUN}} & \multicolumn{2}{c}{\textbf{Places365}} &  \multicolumn{2}{c}{\textbf{Textures}} & \multicolumn{2}{c}{\textbf{Average}}\\
         ~ & FPR95$\downarrow$ & AUROC$\uparrow$ & FPR95$\downarrow$ & AUROC$\uparrow$ & FPR95$\downarrow$ & AUROC$\uparrow$  & FPR95$\downarrow$ & AUROC$\uparrow$  & FPR95$\downarrow$ & AUROC$\uparrow$\\
        \midrule[0.6pt]
        ~ & \multicolumn{10}{c}{1-shot}\\
         LoCoOp & $28.81$ & $94.05$ & $25.76$ & $94.51$ & $33.68$ & $91.59$ & $51.53$ & $86.85$ & $34.94$ & $91.75$ \\
         SCT-L & $19.16$ & $95.70$ & $23.52$ & $94.58$ & $32.81$ & $91.23$ & $48.87$ & $86.66$ & $\textbf{31.09}$ & $92.04$ \\
         SCT-Pow & $18.91$ & $95.84$ & $25.06$ & $94.61$ & $33.36$ & $91.68$ & $49.13$ & $86.68$ & $31.62$ & $\textbf{92.20}$ \\
         SCT-Log & $18.32$ & $95.98$ & $26.00$ & $94.62$ & $33.42$ & $91.39$ & $51.17$ & $85.97$ & $32.23$ & $91.99$ \\
         SCT-Tri & $ 18.46 $ & $ 95.84 $ & $ 24.32 $ & $ 94.98 $ & $ 33.23 $ & $ 91.69 $ & $ 54.10 $ & $ 86.15 $ & $ 32.53 $ & $ 92.17 $ \\
        ~& \multicolumn{10}{c}{16-shot}\\
         LoCoOp & $17.58$ & $96.30$ & $22.82$ & $95.20$ & $32.21$ & $92.03$ & $45.27$ & $88.86$ & $29.47$ & $93.10$ \\
        SCT-L & $13.94$ & $95.86$ & $20.55$ & $95.33$ & $29.86$ & $92.24$ & $41.51$ & $89.06$ & $\textbf{26.47}$ & $\textbf{93.37}$ \\
        SCT-Pow & $14.07$ & $96.70$ & $20.74$ & $94.75$ & $30.11$ & $91.96$ & $43.49$ & $87.90$ & $27.10$ & $92.83$ \\
        SCT-Log & $ 13.11 $ & $ 97.01 $ & $ 20.56 $ & $ 95.50 $ & $ 29.03 $ & $ 92.66 $ & $ 45.55 $ & $ 87.61 $ & $ 27.06 $ & $ 93.20 $ \\
        SCT-Tri & $ 14.88 $ & $ 96.81 $ & $ 20.30 $ & $ 95.44 $ & $ 29.33 $ & $ 92.50 $ & $ 44.84 $ & $ 87.90 $ & $ 27.34 $ & $ 93.16 $ \\
        \bottomrule[1.5pt]
    \end{tabular}}
    \label{tab:exp_instantiation}
\end{table*}

\paragraph{Experiments on conventional CIFAR benchmarks.}

We conduct the experiments on conventional CIFAR benchmarks, following previous works~\citep{liu2020energy}. We adopt CIFAR-10, CIFAR-100\citep{krizhevsky2009learning_cifar10} as the ID datasets and use Textures~\citep{cimpoi2014describing}, Places365~\citep{zhou2016places}, iSUN~\citep{xu2015turkergaze}, LSUN\_Crop~\citep{yu2015lsun}, and LSUN\_Resize~\citep{yu2015lsun}  as the OOD test datasets. We summarize the comparison results averaged across all OOD test datasets in Table~\ref{tab:cifar}, which confirm that SCT achieves superior performance to LoCoOp.

\begin{table}[!ht]
\caption{Experiments on CIFAR benchmark with 16-shot data.}
    \centering
    \small%\renewcommand\arraystretch{0.9}
    \resizebox{0.8\textwidth}{!}{
    \begin{tabular}{lccc|ccc}
        \toprule[1.5pt]
         \multirow{4}*{Method} & \multicolumn{6}{c}{$\mathcal{D}_\text{in}$} \\
         \cmidrule{2-7}
        ~ & \multicolumn{3}{c|}{\textbf{CIFAR-10}} & \multicolumn{3}{c}{\textbf{CIFAR-100}} \\
        ~ & FPR95$\downarrow$ & AUROC$\uparrow$ & ID-ACC$\uparrow$ & FPR95$\downarrow$ & AUROC$\uparrow$ & ID-ACC$\uparrow$ \\
        \midrule[0.6pt]
        MCM & 16.36 & 95.68 & 90.10 & 74.92 & 77.62 & 68.40 \\
        LoCoOp & 15.30 & 95.37 & 93.00 & 64.66 & 82.45 & 72.20\\
        SCT & 12.89 & 96.10 & 93.10 & 60.74 & 84.91 & 72.30 \\
        \bottomrule[1.5pt]
    \end{tabular}}
    \label{tab:cifar}
\end{table}

\paragraph{Experiments on Computational Cost} SCT doesn't incur any extra computational cost to the LoCoOp due to its simple design. Technically, SCT introduces modulating factors respectively on the two components of the original learning objective. The modulating factors (instantiated as $1-p(y|x)$ for $\phi$ and $p(y|x)$ for $\psi$ in Equation (4) in the submission) only involves the computation of the prediction probability of ground truth classes $p(y|x)$, which can be repeatedly used after the original forward pass of CLIP models. What's more, the operation of local features involved in LoCoOp and SCT are also relatively low-cost in terms of computation. The local are generated from the forward pass of the vision encoders of CLIP, which doesn't bring additional computational cost compared to regular training. Regarding the extraction of OOD features, we compute the similarity between every local feature and text feature of all the ID classes and we identify regions that do not include their ground truth class in the top-K predicted classes as ID-irrelevant regions. 
Empirically, we evaluate the time and memory consumption of SCT compared with other baselines in Table~\ref{tab:computational_cost} and the results show that SCT is relatively compute-efficient. The evaluation is conducted on a single GTX-3090 GPU with a batch size as 32.

\begin{table}[!ht]
\caption{Time and memory cost of different methods on the ImageNet benchmark.}
    \centering
    \small%\renewcommand\arraystretch{0.9}
    \resizebox{\textwidth}{!}{
    \begin{tabular}{lccccc}
    \toprule[1.5pt]
        Method & Time for one iteration (s) & GPU Memory (MiB) & FPR95 & AUROC & ID-ACC \\ 
        \midrule[0.6pt]
        CoOp & 0.70 & 21140 & 35.09 & 91.99 & \textbf{71.93} \\ 
        LoCoOp & 0.96 & 23036 & 29.47 & 93.10 & 71.43 \\ 
        SCT & 0.96 & 23036 & \textbf{26.47} & \textbf{93.37} & 71.77 \\ 
        \bottomrule[1.5pt]
    \end{tabular}}
    \label{tab:computational_cost}
\end{table}

\paragraph{Discussions about current advanced post-hoc methods and our method.}

prompt tuning based method can leverage the generalization ability of VLMs to better fit the domains of the downstream tasks with relatively low computational cost. Post-hoc methods need to be built on a well-trained model, the capacity of which greatly affects the OOD detection performance. What's more, post-hoc methods and prompt tuning based methods are compatible with each other, further boosting the OOD detection performance. We conduct experiments on the compatibility of an advanced post-doc method, NegLabel~\citep{jiang2024negative}, with SCT in the Table~\ref{tab:neg_label}, and the results show that SCT can be combined with post-hoc methods for better OOD detection.

In addition, recent advanced post-hoc methods, such ASH~\citep{djurisic2022extremely} and ISH~\citep{xu2023scaling}, improve OOD detection from the perspective of activation scale. However, VLMs like CLIP compute prediction probability based on the cosine similarity between image and text features. The computation of cosine similarity involves the normalization of features, which naturally eliminates the effect of the activation scale. Therefore, ASH and ISH can't apply to VLM models.

\begin{table}[!ht]
\caption{Experiments on compatibility of Neg-Label and SCT on 16-shot data.}
    \centering
    \small%\renewcommand\arraystretch{0.9}
    \resizebox{\textwidth}{!}{
    \begin{tabular}{lllllllllll}
        \toprule[1.5pt]
         \multirow{3}*{\textbf{threshold}} & \multicolumn{8}{c|}{\textbf{OOD dataset}}\\
        %\cmidrule{3-8}
         ~ & \multicolumn{2}{c}{\textbf{iNaturalist}} & \multicolumn{2}{c}{\textbf{SUN}} & \multicolumn{2}{c}{\textbf{Places365}} &  \multicolumn{2}{c}{\textbf{Textures}} & \multicolumn{2}{c}{\textbf{Average}}\\
         ~ & FPR95$\downarrow$ & AUROC$\uparrow$ & FPR95$\downarrow$ & AUROC$\uparrow$ & FPR95$\downarrow$ & AUROC$\uparrow$  & FPR95$\downarrow$ & AUROC$\uparrow$  & FPR95$\downarrow$ & AUROC$\uparrow$\\
        \midrule[0.6pt]
        NegLabel & 1.91 & 99.49 & 20.53 & 95.49 & 35.59 & 91.64 & 43.56 & 90.22 & 25.40 & 94.21 \\
        NegLabel+SCT & 2.03 & 99.51 & 17.42 & 95.92 & 31.43 & 92.46 & 38.46 & 91.23 & 22.34 & 94.78 \\
        \bottomrule[1.5pt]
    \end{tabular}}
    \label{tab:neg_label}
\end{table}

\paragraph{Improvement on confidence calibration.} We use Expected Calibration Error (ECE)~\citep{naeini2015obtaining} to measure the improvement of SCT on confidence calibration over LoCoOp, with lower values indicating better calibration. ECE is calculated by dividing predictions on samples into M equally-spaced bins by confidence scores, then computing the mean average of the difference between each bin’s accuracy and confidence. It can be formulated as $ECE=\sum_{m=1}^M \frac{|Bin_m|}{n}|acc(Bin_m)-conf(Bin_m)|$, where $n$ denotes the number of samples. As shown in Figure~\ref{fig:ece}, SCT can consistently enhance the calibration of VLMs across all few-shot settings.

\subsubsection{More empirical demonstrations and illustrations of our research problem.}
\label{app:more_demo}

In this section, we present more empirical evidence and illustrations of the research problem that inspires our SCT in Fig~\ref{fig:problem}. As illustrated, portions of the extracted local features from ID data are invalid OOD features, thus degrading the performance of OOD detection.

\subsection{Broader impact}
\label{app:impact}
OOD detection is crucial for deploying reliable deep learning systems in real-world scenarios ~\citep{nguyen2015deep,hendrycks2019scaling}. This significance is particularly evident in safety-critical domains such as finance or medical intelligence, where a trustworthy model must accurately distinguish between samples belonging to distinct label spaces (e.g., animals) rather than simply providing predictions based on known classes (e.g., financial products or medical conditions). Our research focuses on a general and practical challenge concerning prompt tuning methods for improving OOD detection efficacy. We specifically focus on the issue of spurious OOD features extracted from ID samples. It is important for effective OOD detection to gain better empirical performance through regularizing prompt tuning.

%%%%%%%%%%%%%%%%%%%%%%%%%%%%%%%%%%%%%%%%%%%%%%%%%%%%%%%%%%%%

\clearpage

\section*{NeurIPS Paper Checklist}

\begin{enumerate}

\item {\bf Claims}
    \item[] Question: Do the main claims made in the abstract and introduction accurately reflect the paper's contributions and scope?
    \item[] Answer: \answerYes{} % Replace by \answerYes{}, \answerNo{}, or \answerNA{}.
    \item[] Justification: The main claims made in the abstract and the introduction accurately reflect the paper's contributions and the scope.

\item {\bf Limitations}
    \item[] Question: Does the paper discuss the limitations of the work performed by the authors?
    \item[] Answer: \answerYes{} % Replace by \answerYes{}, \answerNo{}, or \answerNA{}.
    \item[] Justification: We have discussed the limitations of the work in Section \ref{discuss_limit}.

\item {\bf Theory Assumptions and Proofs}
    \item[] Question: For each theoretical result, does the paper provide the full set of assumptions and a complete (and correct) proof?
    \item[] Answer: \answerNA{} % Replace by \answerYes{}, \answerNo{}, or \answerNA{}.
    \item[] Justification: This paper does not include theoretical results.

    \item {\bf Experimental Result Reproducibility}
    \item[] Question: Does the paper fully disclose all the information needed to reproduce the main experimental results of the paper to the extent that it affects the main claims and/or conclusions of the paper (regardless of whether the code and data are provided or not)?
    \item[] Answer: \answerYes{} % Replace by \answerYes{}, \answerNo{}, or \answerNA{}.
    \item[] Justification: The experimental setups are provided in Section~\ref{sec:setup} to ensure the result reproducibility.

\item {\bf Open access to data and code}
    \item[] Question: Does the paper provide open access to the data and code, with sufficient instructions to faithfully reproduce the main experimental results, as described in supplemental material?
    \item[] Answer: \answerYes{} % Replace by \answerYes{}, \answerNo{}, or \answerNA{}.
    \item[] Justification: The code is publicly available at: \url{https://github.com/tmlr-group/SCT}.

\item {\bf Experimental Setting/Details}
    \item[] Question: Does the paper specify all the training and test details (e.g., data splits, hyperparameters, how they were chosen, type of optimizer, etc.) necessary to understand the results?
    \item[] Answer: \answerYes{} % Replace by \answerYes{}, \answerNo{}, or \answerNA{}.
    \item[] Justification: We provide the detailed experimental setups in Section~\ref{sec:setup} and more discussion in Appendix~\ref{app:baseline_info}.

\item {\bf Experiment Statistical Significance}
    \item[] Question: Does the paper report error bars suitably and correctly defined or other appropriate information about the statistical significance of the experiments?
    \item[] Answer: \answerYes{} % Replace by \answerYes{}, \answerNo{}, or \answerNA{}.
    \item[] Justification: We demonstrate the experimental statistical significance by reporting the mean and std value based on multiple trials of the experiments in Tables~\ref{tab:main_table1},~\ref{tab:main_table2}, and other results.

\item {\bf Experiments Compute Resources}
    \item[] Question: For each experiment, does the paper provide sufficient information on the computer resources (type of compute workers, memory, time of execution) needed to reproduce the experiments?
    \item[] Answer: \answerYes{} % Replace by \answerYes{}, \answerNo{}, or \answerNA{}.
    \item[] Justification: We provide the details of the used experiment compute resources in the reproducibility statement in Appendix.
    
\item {\bf Code Of Ethics}
    \item[] Question: Does the research conducted in the paper conform, in every respect, with the NeurIPS Code of Ethics \url{https://neurips.cc/public/EthicsGuidelines}?
    \item[] Answer: \answerYes{} % Replace by \answerYes{}, \answerNo{}, or \answerNA{}.
    \item[] Justification: The research conducted in the paper conforms, in every respect, with the NeurIPS Code of Ethics.

\item {\bf Broader Impacts}
    \item[] Question: Does the paper discuss both potential positive societal impacts and negative societal impacts of the work performed?
    \item[] Answer: \answerYes{} % Replace by \answerYes{}, \answerNo{}, or \answerNA{}.
    \item[] Justification: We discuss the potential positive and negative societal impacts of the work in Appendix~\ref{app:impact}.
    
\item {\bf Safeguards}
    \item[] Question: Does the paper describe safeguards that have been put in place for responsible release of data or models that have a high risk for misuse (e.g., pretrained language models, image generators, or scraped datasets)?
    \item[] Answer: \answerNA{} % Replace by \answerYes{}, \answerNo{}, or \answerNA{}.
    \item[] Justification: This paper poses no such risks.

\item {\bf Licenses for existing assets}
    \item[] Question: Are the creators or original owners of assets (e.g., code, data, models), used in the paper, properly credited and are the license and terms of use explicitly mentioned and properly respected?
    \item[] Answer: \answerYes{} % Replace by \answerYes{}, \answerNo{}, or \answerNA{}.
    \item[] Justification: The creators or original owners of the datasets and referred codes are properly mentioned in the references and their credits are properly respected.

\item {\bf New Assets}
    \item[] Question: Are new assets introduced in the paper well documented and is the documentation provided alongside the assets?
    \item[] Answer: \answerNA{} % Replace by \answerYes{}, \answerNo{}, or \answerNA{}.
    \item[] Justification: This paper does not release new assets.

\item {\bf Crowdsourcing and Research with Human Subjects}
    \item[] Question: For crowdsourcing experiments and research with human subjects, does the paper include the full text of instructions given to participants and screenshots, if applicable, as well as details about compensation (if any)? 
    \item[] Answer: \answerNA{} % Replace by \answerYes{}, \answerNo{}, or \answerNA{}.
    \item[] Justification: This paper does not involve crowdsourcing nor research with human subjects.

\item {\bf Institutional Review Board (IRB) Approvals or Equivalent for Research with Human Subjects}
    \item[] Question: Does the paper describe potential risks incurred by study participants, whether such risks were disclosed to the subjects, and whether Institutional Review Board (IRB) approvals (or an equivalent approval/review based on the requirements of your country or institution) were obtained?
    \item[] Answer: \answerNA{} % Replace by \answerYes{}, \answerNo{}, or \answerNA{}.
    \item[] Justification: This paper does not involve crowdsourcing nor research with human subjects.

\end{enumerate}

\appendix

\end{document}